%% file: _main.tex
\documentclass[twoside]{article}

\usepackage[accepted]{aistats2020}

\usepackage[ruled,vlined]{algorithm2e}
\DontPrintSemicolon

\usepackage[round,sort&compress]{natbib}

\usepackage{changepage}
\usepackage{booktabs}
\RequirePackage[colorlinks,citecolor=blue,urlcolor=blue]{hyperref}
\RequirePackage[usenames,dvipsnames]{color}
\usepackage{url}
\usepackage{times}
\usepackage{bm}
\usepackage{mathrsfs}
\usepackage{graphicx}
\usepackage{subcaption}
\usepackage{floatrow}
\usepackage{enumitem}
\usepackage{jhh-misc2}
\crefname{lemma}{Lemma}{Lemmas}
\crefname{proposition}{Proposition}{Propositions}
\crefname{corollary}{Corollary}{Corollaries}
\crefname{theorem}{Theorem}{Theorems}
\crefname{example}{Example}{Examples}
\crefname{assumption}{Assumption}{Assumptions}
\usepackage{autonum}

\input{shortcuts}

\graphicspath{ {figures/} }

\newfloatcommand{capbtabbox}{table}[][\FBwidth]
\begin{document}

\twocolumn[

\aistatstitle{Validated Variational Inference via Practical Posterior Error Bounds}

\aistatsauthor{Jonathan H.~Huggins \And Miko{\l}aj Kasprzak \And Trevor Campbell \And Tamara Broderick}

\aistatsaddress{Boston University \And University of Luxembourg \And  University of British Columbia \And MIT}
]

\input{abstract}
\input{introduction}

\input{prelim}
\input{discrepancybounds}

\input{computablebounds}
\input{experiments}

\input{discussion} %

\bibliographystyle{abbrvnat}
\bibliography{library,library_extra}

\input{appendix}

\end{document}

%% file: shortcuts.tex
\newcommand{\WHsimple}[2]{\mathrm{W}_{#1}\mathrm{H}(#2)}

\newcommand{\EI}[2]{\mathrm{EI}_{#1}(#2)}
\newcommand{\EIconst}[3]{\mathrm{EI}_{#1}^{*}(#2, #3)}

\renewcommand{\Pr}{\mathbb{P}}
\newcommand{\metric}{m}

\newcommand{\elbo}[1]{\ensuremath{\mathsf{ELBO}(#1)}\xspace}
\newcommand{\cubo}[2]{\ensuremath{\mathsf{CUBO}_{#1}(#2)}\xspace}
\newcommand{\PIC}[2]{C^\mathrm{PI}_{#1}(#2)}
\newcommand{\EIC}[2]{C^\mathrm{EI}_{#1}(#2)}
\newcommand{\divbd}[3]{\mathrm{H}_{#1}(#2, #3)}

\newcommand{\elboEst}[1]{\ensuremath{\widehat{\mathsf{ELBO}}(#1)}\xspace}
\newcommand{\cuboEst}[2]{\ensuremath{\widehat{\mathsf{CUBO}}_{#1}(#2)}\xspace}

\newcommand{\EICest}[2]{\widehat C^\mathrm{EI}_{#1}(#2)}
\newcommand{\divbdEst}[3]{\widehat{\mathrm{H}}_{#1}(#2, #3)}

\newcommand{\distWeibull}{\distNamed{Weibull}}

\newcommand{\targetdist}{\pi}
\newcommand{\postdist}{\targetdist}
\newcommand{\approxdist}{\hat{\targetdist}}
\newcommand{\vardist}{\xi}
\newcommand{\varfamily}{\mcQ}
\newcommand{\lik}{\ell}

\newcommand{\marginallik}{M}

\newcommand{\discrepancy}[2]{{\mcD_{#1}(#2)}}

\newcommand{\mainmeas}{\nu}
\newcommand{\estmeas}{\eta}

\newcommand{\Xvec}{X}

\newcommand{\priordist}{\targetdist_{0}}

\newcommand{\param}{\theta}
\newcommand{\paramrv}{\vartheta}

\newcommand{\abscont}{\ll}  

\newcommand{\couplings}[2]{\Gamma(#1, #2)}
\newcommand{\coupling}{\gamma}

\newcommand{\pwassSimple}[3]{\mcW_{#1}(#2, #3)}

\newcommand{\MAD}[1]{\operatorname{MAD}_{#1}}

\newcommand{\mean}[1]{m_{#1}}
\newcommand{\meanvec}[1]{m_{#1}}

\newcommand{\var}[1]{\sigma_{#1}^{2}}
\newcommand{\std}[1]{\sigma_{#1}}

\def\norm#1{\left\|{#1}\right\|} 

\newcommand{\twonorm}[1]{\norm{#1}_2} 
\def\staticnorm#1{\|{#1}\|} 
\newcommand{\statictwonorm}[1]{\staticnorm{#1}_2} 

\def\Renyi{R\'enyi\xspace}

%% file: abstract.tex
\begin{abstract}
Variational inference has become an increasingly attractive fast alternative to 
Markov chain Monte Carlo methods for approximate Bayesian inference.
However, a major obstacle to the widespread use of variational methods is the lack of post-hoc accuracy
measures that are both theoretically justified and computationally efficient.
In this paper, we provide rigorous bounds on the error of posterior mean and uncertainty estimates that arise from full-distribution approximations, as in variational inference.
Our bounds are widely applicable, as they require only that the approximating and exact posteriors have polynomial moments.
Our bounds are also computationally efficient for variational inference because they require only standard values from variational objectives, 
straightforward analytic calculations, and simple Monte Carlo estimates.
We show that our analysis naturally leads to a new and improved workflow for validated variational inference.
Finally, we demonstrate the utility of our proposed workflow and error bounds on a robust regression problem
and on a real-data example with a widely used multilevel hierarchical model.

\end{abstract}

%% file: introduction.tex
\section{Introduction}

Exact Bayesian statistical inference is known for providing point estimates with desirable decision-theoretic properties as well as coherent uncertainties. 
Using Bayesian methods in practice, though, typically requires approximating these quantities. 
Therefore, it is crucial to quantify the error introduced by any approximation. 
There are two, essentially complementary, options: (1) rigorous \emph{a priori} characterization of accuracy for finite data and (2) tools for evaluating approximation accuracy \emph{a posteriori}.
First, consider option \#1.
Markov chain Monte Carlo (MCMC) methods are the gold standard for sound approximate Bayesian inference in part due to their flexibility and 
strong \emph{a priori} theoretical guarantees on quality for finite data. However, these guarantees are typically asymptotic in running time, and computational concerns have motivated a spate of alternative Bayesian approximations. Within the machine learning community, variational approaches \citep{Blei:2017,wainwright2008graphical} such as black-box and automatic differentiation variational inference \citep{Ranganath:2014,Kucukelbir:2015,Kingma:2014}
are perhaps the most widely used. 
While these methods have empirically demonstrated computational gains on problems of interest, they do not come equipped with guarantees on the 
approximation accuracy of point estimates and uncertainties.
There has been some limited but ongoing work in developing relevant \emph{a priori} guarantees for common variational approaches \citep{Alquier:2016b,Alquier:2017,CheriefAbdellatif:2018,Wang:2018,Pati:2018,Wang:2019:VB-misspecified}. 
There has also been work in developing (boosting) variational algorithms for which it may be possible to obtain \emph{a priori} guarantees on
convergence of the approximating distribution to arbitrary accuracy~\citep{Guo16,Wang16,Miller17,Locatello18,Locatello18b,Campbell19}.

The examples above typically either have no guarantees or purely asymptotic guarantees -- or require non-convex optimization. 
Thus, in every case, reliable evaluation tools (option \#2) would provide an important bulwark for data analysis
(as demonstrated by the widespread use of convergence diagnostics for MCMC \citep{Gelman:2013}). 
In any particular data analysis, such tools could determine if the approximate point estimates and uncertainties are to be trusted.
\citet{Yao:2018,Gorham:2015,Gorham:2017,Gorham:2019} have pioneered initial work in developing evaluation tools applicable to variational inference. 
However, current methods are either heuristic or cannot be applied in an automated way. %

In this paper, we provide the first rigorous, automated, \emph{and} computationally efficient error bounds on the quality of posterior point and uncertainty estimates for variational approximations. We highlight three practical aspects of our bounds here: (A) computational efficiency, (B) weak tail restrictions, and (C) relevant targets.
For A, we use only standard values computed in the course of variational inference, straightforward analytic calculations, and simple Monte Carlo (not MCMC) estimates.
For B, we require only that the approximating and exact posteriors have polynomial moments -- though we show even tighter bounds when exponential moments exist.
For C, note that practitioners typically report posterior means for point estimates -- and they report posterior variance,  
standard deviation, or mean absolute deviation for uncertainties \citep{Robert:1994,Gelman:2013}. So we directly bound the error in these quantities. We demonstrate the importance of bounding error in these output quantities directly, rather than bounding divergences between distributions, with illustrative counterexamples; namely, we show that common variational objectives such as the Kullback--Leibler (KL) divergence and $\alpha$-divergences can be very small at the same time that mean and variance estimates are \emph{arbitrarily} wrong.

To obtain our bounds, we make three main technical contributions, which may all be of independent interest beyond Bayesian methods.
First, we show how to bound mean and uncertainty differences in terms of Wasserstein distance.
Second, we develop novel bounds on the Wasserstein distance in terms of $\alpha$-divergences -- including the KL divergence  -- and 
moment bounds on the variational approximation. 
The moment conditions allow us to relate (scale-free) $\alpha$-divergences to (scale-sensitive) Wasserstein distances.
Finally, we derive efficiently computable bounds on $\alpha$-divergences in terms of the objectives already widely used for variational inference
-- in particular, the evidence lower bound (ELBO) and $\chi$ upper bound (CUBO) \citep{Dieng:2017}.
By combining all three contributions, we obtain efficiently computed bounds on means and uncertainties in terms of the ELBO, CUBO, 
and certain polynomial or exponential moments of the variational approximation.
	
Our methods give rise to a new and improved workflow for validated variational inference. 
We illustrate the usefulness of our bounds as well as the practicality of our new workflow on a toy robust regression problem 
and a real-data example with a widely used multilevel hierarchical model.
A \texttt{python} package for carrying out our workflow -- including doing black-box variational inference and computing the bounds we develop in this paper --
 is available at \url{https://github.com/jhuggins/viabel}. 
The same repository also contains  code for reproducing all of our experiments. 
Proofs of all our results are in \cref{sec:proofs}.

%% file: prelim.tex
\section{Preliminaries} 
\label{sec:preliminaries}

\textbf{Bayesian inference.} 
Let $\param \in \reals^{d}$ denote a parameter vector of interest, and let $z$
denote observed data. A Bayesian model consists of a prior measure
$\priordist(\dee\param)$ and a likelihood $\lik(z; \param)$.  Together, the
prior and likelihood define a joint distribution over the data and parameters.
The Bayesian posterior distribution $\postdist$ is the conditional in $\param$
with fixed data $z$.\footnote{Since the data $z$ are always fixed throughout
this work, we have suppressed the dependence on $z$ in the notation.} To write
this conditional, we define the unnormalized posterior measure
$\postdist^{*}(\dee\param) \defined \lik(z;\param)\priordist(\dee\param)$ and the
marginal likelihood, or evidence, $\marginallik \defined \int \dee\postdist^{*}$.
Then the posterior is
$
\postdist \defined \postdist^{*}/M. %
$

Typically, practitioners report \emph{summaries} -- e.g., point estimates and uncertainties -- of the posterior rather than the full posterior. 
Such summaries include the mean $\mean{\postdist}$, covariance $\Sigma_{\postdist}$, $i$th component marginal standard deviation $\std{\postdist,i}$, and
mean absolute deviation $\MAD{\postdist,i}$: for $\paramrv \dist \postdist$, %
\[
\mean{\postdist} &\defined \EE(\paramrv), & 
\MAD{\postdist,i} &\defined \EE(|\paramrv_{i} - \mean{\postdist,i}|), \\
\std{\postdist,i} &\defined \Sigma_{\approxdist,ii}^{1/2}, & 
\Sigma_{\postdist} &\defined \EE\{(\paramrv - \meanvec{\postdist})(\paramrv - \meanvec{\postdist})^{\top}\}.
\]

\textbf{Variational inference.}
In most applications of interest, it is infeasible to efficiently compute
these summaries with respect to the posterior distribution in closed form or via
simple Monte Carlo.  Therefore, one must use an approximate inference method,
which produces an approximation $\approxdist$ to the posterior
$\postdist$.  
The summaries of $\approxdist$ may in turn be used as
approximations to the summaries of $\postdist$. 
One approach, \emph{variational inference}, is widely used in machine learning.
Variational inference aims to minimize some
measure of discrepancy $\discrepancy{\postdist}{\cdot}$ over a tractable family
$\varfamily$ of potential approximation
distributions~\citep{wainwright2008graphical,Blei:2017}:
\[
\approxdist = \argmin_{\vardist \in \varfamily} \discrepancy{\postdist}{\vardist}.
\]
The variational family $\varfamily$ is chosen to be tractable in the sense
that, for any $\vardist \in \varfamily$, we are able to efficiently calculate
relevant summaries either analytically or using independent and identically
distributed samples from $\vardist$.  

\textbf{KL divergence.} 
The classical choice for the discrepancy in variational inference is the
\emph{Kullback--Leibler (KL) divergence} (or \emph{relative entropy}) \citep{bishop2006pattern}:
\[
\kl{\vardist}{\postdist} \defined \int \log \left(\der{\vardist}{\postdist}\right) \dee\vardist.
\]
Note that the KL divergence is asymmetric in its arguments. The direction
$\discrepancy{\postdist}{\vardist} = \kl{\vardist}{\postdist}$ is most typical in variational inference, largely out of
convenience; the unknown marginal likelihood $M$ appears in an additive constant that does not influence
the optimization, and computing gradients requires estimating expectations
only with respect to $\vardist \in \varfamily$, which is chosen to be tractable. %
Minimizing $\kl{\vardist}{\postdist}$ is equivalent to maximizing 
the \emph{evidence lower bound} \citep[ELBO;][]{bishop2006pattern}: %
\[
\elbo{\vardist} &\defined  \int \log\left(\der{\postdist^{*}}{\vardist} \right)\dee\vardist. %
\]

\textbf{\Renyi's $\alpha$-divergence.}
Another choice of discrepancy for variational inference \citep{HernandezLobato:2016,Li:2016,Bui:2017a,Dieng:2017} is \emph{\Renyi's $\alpha$-divergence}, 
which for $\alpha \in (0,1) \cup (1,\infty)$ is defined as
\[
\alphaDiv{\alpha}{\postdist}{\vardist} &\defined \frac{1}{\alpha-1}\log\int \left(\der{\postdist}{\vardist}\right)^{\alpha-1} \dee\postdist.
\]
The $\alpha$-divergence is typically used in variational inference with $\discrepancy{\postdist}{\vardist} =
\alphaDiv{\alpha}{\postdist}{\vardist}$ for $\alpha > 1$; again, the unknown marginal likelihood $M$ 
does not influence the optimization, and estimating gradients is tractable. 
Variational inference with the $\alpha$-divergence is equivalent to
minimizing a quantity known as the \emph{$\chi$ upper bound} \citep[CUBO;][]{Dieng:2017}:
\[
\cubo{\alpha}{\vardist} \defined (1 - \alpha^{-1})\alphaDiv{\alpha}{\postdist}{\vardist} - \log \marginallik.
\]
The ELBO and CUBO are so-named since they respectively provide a lower and upper bound for $\log M$; see \cref{sec:elbo_cubo_bounds_proof}.
The $\alpha$-divergence  generalizes the KL divergence since 
$\alphaDiv{\alpha}{\postdist}{\vardist}  \defined \lim_{\alpha \to 1} \alphaDiv{\alpha}{\postdist}{\vardist} = \kl{\postdist}{\vardist}$ \citep{Cichocki:2010}.
Note, however, that here the KL divergence has the order of its arguments switched when compared to
how it is used for variational inference.

\textbf{Wasserstein distance.}
The Wasserstein distance is a measure of discrepancy 
that, unlike the previous two divergences, is influenced by
a metric on the space on which the distributions are defined.  
It is widely used in the analysis of MCMC and large-scale data 
asymptotics~\citep[e.g.,][]{Joulin:2010,Madras:2010,Rudolf:2015,Durmus:2019,Durmus:2019b,Vollmer:2016,Eberle:2018}. %
The $p$-Wasserstein distance
between $\vardist$ and $\postdist$ is given by
\[
\pwassSimple{p}{\vardist}{\postdist} 
\defined \inf_{\coupling \in \couplings{\vardist}{\postdist}} \left\{ \int \staticnorm{\param - \param'}_{2}^{p} \coupling(\dee \param, \dee \param') \right\}^{1/p},
\]
where $\couplings{\vardist}{\postdist}$ is
the set of \emph{couplings} between $\vardist$ and $\postdist$, i.e.,
Borel measures $\coupling$ on
$\reals^{d} \times \reals^{d}$ such that $\vardist = \coupling(\cdot, \reals^{d})$ and
$\postdist = \coupling(\reals^{d}, \cdot)$~\citep[Defs.~6.1 \& 1.1]{Villani:2009}. 
The Wasserstein distance is difficult to use as a variational objective 
due to the (generally intractable) infimum over couplings, although there is
recent work in this direction \citep{cuturi2014fast,claici2018stochastic,srivastava2018scalable}.

\subsection{Previous work on validating variational approximations} \label{sec:previous-work}

\textbf{Stein discrepancies.} 
Computable Stein discrepancies provide one approach for evaluating variational approximations \citep{Gorham:2015,Gorham:2017,Gorham:2019}.
That is, the Stein discrepancy between the posterior and variational approximation could be approximated using samples from the variational approximation. 
However, Stein discrepancy-based bounds on the Wasserstein distance require knowledge of certain properties of the posterior (e.g., one-sided Lipschitz constants) 
that are usually unavailable without additional analytic effort.
Thus, there is not yet an automated way to apply an appropriate Stein operator that guarantees control 
of the Wasserstein distance~\citep{Erdogdu:2018,Gorham:2019}.

\textbf{Pareto-smoothed importance sampling and $\hat{k}$.}
Pareto-smoothed importance sampling~\citep[PSIS;][]{Vehtari:2019} is a method for reducing the variance of importance sampling estimators. 
The key quantity computed in PSIS is $\hat{k}$, which is an estimate of $k \defined \inf \{ k' \mid \alphaDiv{1/k'}{\targetdist}{\approxdist} < \infty \}$. 
\citet{Yao:2018} suggest using $\hat{k}$ as a measure of the quality of $\approxdist$. 
Based on the empirical results and informal arguments of \citet{Vehtari:2019}, they propose that $\hat{k} \le 0.5$ indicates a good variational approximation
and $\hat{k} \in [0.5, 0.7]$ indicates minimal acceptability. In all cases the authors suggest using PSIS to improve estimates of posterior expectations. 
However, the link between a small $\hat{k}$ value and a high-quality posterior approximation is only heuristic.
We find empirically in \cref{sec:experiments,sec:robust-regression} that poor posterior approximations can have small $\hat{k}$ values. 

%% file: discrepancybounds.tex
\section{Error bounds via posterior discrepancies}
\label{sec:problem}

Given the concern with posterior summaries, a meaningful measure of posterior approximation quality should
 control the error in each of these summaries, i.e.,
$\|\mean{\approxdist} - \mean{\postdist}\|_2$, 
$|\MAD{\approxdist,i} - \MAD{\postdist,i}|$, $\|\Sigma_{\approxdist} - \Sigma_\postdist\|_2$, and
$|\std{\approxdist,i} - \std{\postdist,i}|$. %
To be practical, this measure should also be computationally efficient.
We start by focusing on the former challenge: finding a discrepancy that controls the error of these summaries. 
In particular, we (1) provide counterexamples to show that $\kl{\approxdist}{\postdist}$ and $\alphaDiv{\alpha}{\postdist}{\approxdist}$
by themselves cannot be relied upon to control these errors, and (2) prove that the Wasserstein distance does provide
the desired control.
We address the latter challenge, computational efficiency, in \cref{sec:worflow}.

\textbf{KL divergence.} %
Unfortunately, as we show in the following examples, %
even when $\kl{\approxdist}{\postdist}$ is small, posterior summary approximations provided by $\approxdist$ can be arbitrarily poor. 
To get a sense of scale for the KL divergence, we note that the KL divergence from a variational approximation to the exact posterior 
can easily range from 1 to nearly 500. See \cref{sec:KL-discussion} for further discussion. 
First we note that the exact posterior standard deviation $\std{\postdist}$ is a natural scale for the posterior mean error since changing the posterior mean by $\std{\postdist}$ or more could fundamentally change practical decisions made based on the posterior.
Our first example shows that even when $\kl{\approxdist}{\postdist}$ is small, the mean error can be arbitrarily large, 
whether measured relative to $\std{\postdist}$ or $\std{\approxdist}$.
\begin{proposition}[Arbitrarily poor mean approximation]\label{ex:KL-divergence-problems1}
For any $t>0$, there exist
(A) one-dimensional, unimodal distributions $\approxdist$ and $\postdist$ such that $\kl{\approxdist}{\postdist}< 0.9$ and $(\mean{\approxdist}-\mean{\postdist})^2>t\var{\postdist}$,
and (B) one-dimensional, unimodal distributions $\approxdist$ and $\postdist$ such that $\kl{{\approxdist}}{{\postdist}}< 0.3$ and $(\mean{{\approxdist}}-\mean{{\postdist}})^2>t\var{\approxdist}.$
\end{proposition}
In more detail, let $\distWeibull(k, 1)$ denote the Weibull distribution with shape $k>0$ and scale 1.
For (A), for any $t > 0$, we can choose $k = k(t)$, $\postdist = \distWeibull(k, 1)$, and $\approxdist = \distWeibull(k/2, 1)$, where $k(t) \searrow 0$ as $t \to \infty$. 
We exchange the two distributions for (B).

Our second example shows that $\kl{\approxdist}{\postdist}$ can remain small even when the variance difference is arbitrarily large.%
\begin{proposition}[Arbitrarily poor variance approximation]\label{ex:KL-divergence-problems2}
For any $t \in (1, \infty]$, there exist one-dimensional, mean-zero, unimodal distributions $\approxdist$ and $\postdist$ such that 
${\kl{\approxdist}{\postdist}} < 0.12$ but $\var{\postdist} \ge t \var{\approxdist}$. 
\end{proposition}
Here, for any $t > 0$ we let $h=h(t)$, $\postdist = \mcT_{h}$ (standard $t$-distribution with $h$ degrees of freedom), 
and $\approxdist = \distNorm(0,1)$ (standard Gaussian), where $h(t) \searrow 2$ as $t \to \infty$.

\textbf{\Renyi's $\alpha$-divergence.} %
We similarly demonstrate that small $\alphaDiv{\alpha}{\postdist}{\approxdist}$ does not imply accurate mean or variance estimates. 
We focus on the canonical case $\alpha=2$, which will also play a key role in our analyses below.
\begin{proposition}[Arbitrarily poor mean and variance approximation] \label{ex:2-div-problems}
For any $t>0$, there exist one-dimensional, unimodal distributions $\approxdist$ and $\postdist$ with ${\alphaDiv{2}{\postdist}{\approxdist}} < 0.4$
 such that $\sigma_{\approxdist}^2\ge t\sigma_{\postdist}^2$ and $(\mean{{\approxdist}}-\mean{{\postdist}})^2\ge t\var{\postdist}$.
\end{proposition}
We again take $\postdist = \distWeibull(k, 1)$ and $\approxdist = \distWeibull(k/2, 1)$ with $k = k(t) \searrow 0$ as $t \to \infty$.

\textbf{Wasserstein distance.}
In contrast to both the KL and $\alpha$-divergences, the Wasserstein distance 
accounts for the metric on the underlying space. 
Intuitively, the Wasserstein distance is large when the mass of two distributions is ``far apart.''
Thus, it is a natural choice of discrepancy for bounding
the error in the approximate posterior mean and uncertainty, since these quantities also 
depend on the underlying metric. 
Our next result confirms that the Wasserstein
distance controls the error in these quantities. %
\begin{theorem} \label{thm:Wasserstein-moment-bounds}
If $\pwassSimple{1}{\approxdist}{\postdist} \le \veps$ or $\pwassSimple{2}{\approxdist}{\postdist} \le \veps$, then 
\[
\twonorm{\mean{\approxdist} - \mean{\postdist}} \le \veps 
\,\,\, \text{and}\,\,\,
\max_i|\MAD{\approxdist,i} - \MAD{\postdist,i}| \le 2\veps.
\]
If $\pwassSimple{2}{\approxdist}{\postdist} \le \veps$, then, for $S \defined \sqrt{\min\{\twonorm{\Sigma_{\approxdist}},\twonorm{\Sigma_{\postdist}}\}}$,
\[
\max_i |\std{\approxdist,i} - \std{\postdist,i}| \le \veps 
\,\,\, \text{and}\,\,\,
\twonorm{\Sigma_{\approxdist} - \Sigma_{\postdist}} < 2\veps(S + \veps). %
\]
\end{theorem}

\begin{remark}
The Wasserstein distance can also be used to bound the difference between expectations of any smooth function.
More precisely, if the function $\phi$ satisfies $|\phi'| \le L$ and $\pwassSimple{1}{\approxdist}{\postdist} \le \veps$, 
then $|\int \phi\,\dee\approxdist - \int \phi\,\dee\postdist| \le \veps L$. 
\end{remark}

While our focus is on parameter inference, there are many cases when we are interested in \emph{predictive accuracy}, including in Bayesian deep learning.
In such cases, Wasserstein bounds on the posterior remain useful.
Assuming $f(z_\text{new} \given \param)$  is the distribution for new data given parameter $\param$, the posterior predictive distribution 
is $\mu(z_\text{new}) \defined \int f(z_\text{new} \given \param) \postdist(\dee\param)$, with the approximate posterior predictive $\hat{\mu}$ 
defined analogously. 
\begin{proposition} \label{prop:predictive-Wasserstein}
If $\pwassSimple{p}{f(\cdot \given \param)}{f(\cdot \given \param')} \le C \twonorm{\param - \param'}$ for some $C \ge 0$
and $\pwassSimple{p}{\approxdist}{\postdist} \le \veps$, then $\pwassSimple{p}{\hat{\mu}}{\mu} \le \veps C$.
\end{proposition}
\begin{remark}
The assumption on $f(\cdot \given \param)$ in \cref{prop:predictive-Wasserstein} holds for many commonly used distributions including Gaussian distributions with fixed variance 
and the Bernoulli distribution with softmax parameterization. 
\end{remark}

%% file: computablebounds.tex
\section{A complete workflow for validated variational inference} \label{sec:worflow}

In this section, we develop a comprehensive approach to variational inference with
rigorously validated output.
First we prove a number of new results which let us bound the Wasserstein distance between the variational
and true posteriors in terms of quantities that can be efficiently computed or upper-bounded. %
These Wasserstein bounds, when combined with ideas from importance sampling, provide the tools 
for formulating our proposed workflow. 

\subsection{Computationally efficient %
error bounds}
\label{sec:computablebounds}

In this section we return to the question of computationally efficient posterior error bounds.
In particular, although we have shown that the Wasserstein distance provides direct control of the error in approximate posterior summaries
of interest, it itself is not tractable to compute or estimate.
Our general strategy in this section is use standard variational objectives -- namely, 
the ELBO and CUBO -- to bound the Wasserstein distance. 
We thereby achieve bounds on the error of posterior summaries by \cref{thm:Wasserstein-moment-bounds}. 
More detail about our results in this section and related work can be found in, respectively, \cref{sec:intuition-and-details,sec:more-transport-entropy}.

Our process consists of two steps.
First, we use tail properties of the distribution $\vardist\in\varfamily$
to arrive at bounds on the Wasserstein distance via the KL or $\alpha$-divergence.
Second, we use ELBO and CUBO to bound
the KL and $\alpha$-divergences. 

A key challenge in realizing our goal is bounding a scale-dependent distance (the Wasserstein distance)
with a scale-invariant divergence (the KL or $\alpha$-divergence).
To see the scale-invariance, %
we note a broader result: these divergences are invariant to
reparameterization.  For a transformation $T : \reals^{d} \to \reals^{d}$, let
$T \# \estmeas$ denote the pushforward measure of $\eta$, which is the distribution of
the random variable $T(\paramrv)$ for $\paramrv \dist \estmeas$. 
\begin{lemma}\label{lem:scale-invariance}
The KL and $\alpha$-divergence are invariant under a smooth, invertible transformation $T$, i.e.,
$\alphaDiv{\alpha}{\estmeas}{\mainmeas} = \alphaDiv{\alpha}{T \# \estmeas}{T \# \mainmeas}$
and $\kl{\estmeas}{\mainmeas} = \kl{T\#\estmeas}{T\#\mainmeas}$.
\end{lemma}
\citet{Yao:2018} make a similar observation.  
A simple example illustrates that the Wasserstein distance is not
invariant to reparameterization. For $\sigma > 0$, let
$\mainmeas_{\sigma}(\param)  \defined \sigma^{-d} \mainmeas(\param/\sigma)$
define the rescaled version of $\mainmeas$ (with $\estmeas_{\sigma}$ defined
analogously).  Then
$\pwassSimple{p}{\estmeas_{\sigma}}{\mainmeas_{\sigma}} = \sigma
\pwassSimple{p}{\estmeas}{\mainmeas}$. 
It follows that any bound of a scale-dependent distance such as the Wasserstein distance
using a scale-invariant divergence must incorporate some notion of scale.
Toward that end, we start by defining the moment constants $\PIC{p}{\vardist}$ and $\EIC{p}{\vardist}$ and associated tail behaviors.
For $p\geq 1$, we say that $\vardist$ is \emph{$p$-polynomially integrable} if 
\[
\PIC{p}{\vardist} \defined 2 \inf_{\param_0}\textstyle\left\{\int \staticnorm{\param - \param_0}_{2}^{p}\vardist(\dee \param)\right\}^{\frac{1}{p}} < \infty \label{eq:polynomialC}
\]
and that $\vardist$ is \emph{$p$-exponentially integrable} if
\[
\EIC{p}{\vardist} \defined 2\!\!\! \inf_{\theta_0, \eps > 0}\textstyle\left[\frac{1}{\eps}
\!\left\{ \frac{3}{2} + \log\int e^{\eps\staticnorm{\param - \param_0}_{2}^{p}}\vardist(\dee \param)\right\}\right]^{\frac{1}{p}}\!\! < \infty.
\]
Assuming the variational approximation $\approxdist$ has polynomial (respectively, exponential) tails,
our next result provides a bound on the $p$-Wasserstein distance using the $2$-divergence (respectively, the KL divergence).
\begin{proposition}\label{thm:Wasserstein-simple}
If $\postdist \abscont \approxdist$,\footnote{$\postdist \abscont \approxdist$ denotes $\postdist$ is absolutely continuous with respect to $\approxdist$.} then
\[
\pwassSimple{p}{\approxdist}{\postdist} &\leq \PIC{2p}{\approxdist} \left[\exp\left\{\alphaDiv{2}{\postdist}{\approxdist}\right\} - 1\right]^{\frac{1}{2p}}
\]
and
\[
\pwassSimple{p}{\approxdist}{\postdist} &\le \EIC{p}{\approxdist}\left[\kl{\postdist}{\approxdist}^{\frac{1}{p}} + \{\kl{\postdist}{\approxdist}/2\}^{\frac{1}{2p}}\right]. 
\]
\end{proposition}
Our next results confirm that, even though \cref{thm:Wasserstein-simple} uses
KL and $\alpha$-divergences, our bounds capture the growth in Wasserstein distance, as desired, and 
thus do not suffer the pathologies observed in \cref{ex:KL-divergence-problems1,ex:KL-divergence-problems2,ex:2-div-problems}.
\begin{proposition}[cf.\ \cref{ex:KL-divergence-problems1,ex:2-div-problems}] \label{ex:no-problems1}
For a fixed $k\in (0,\infty)$, let $\estmeas = \distWeibull(k/2, 1)$ and $\mainmeas = \distWeibull(k, 1)$. %
Then, for $\alpha>1$, $\alphaDiv{\alpha}{\estmeas}{\mainmeas} = \infty$.
On the other hand, $\alphaDiv{\alpha}{\mainmeas}{\estmeas} < \infty$; but, as $k \searrow 0$, 
the moment constant from \cref{thm:polynomial-moments-simple} satisfies $\PIC{p}{\estmeas} \nearrow \infty$. 
\end{proposition}
\begin{proposition}[cf.\ \cref{ex:KL-divergence-problems2}] \label{ex:no-problems2}
If $\estmeas$ is a standard normal measure and $\mainmeas= \mcT_{h}$ is a standard $t$-distribution 
with $h\geq 2$ degrees of freedom, then $\alphaDiv{\alpha}{\estmeas}{\mainmeas}<\infty$.
However, as $h\searrow 2$, we have $\PIC{p}{\mainmeas} \nearrow \infty$. %
\end{proposition}

Next, we turn to showing how we can use the ELBO and CUBO to bound
the KL and $\alpha$-divergences that appear in \cref{thm:Wasserstein-simple}.
For $\alpha > 1$ and any distribution $\estmeas$, define
\[
\divbd{\alpha}{\vardist}{\eta} \defined \textstyle{\frac{\alpha}{\alpha-1}}\left\{\cubo{\alpha}{\vardist} - \elbo{\eta}\right\}.
\]
\begin{lemma} \label{lem:divergence-elbo-cubo-bounds}
For any distribution $\estmeas$ such that $\postdist \abscont \estmeas$,
\[
\kl{\postdist}{\approxdist}\leq \alphaDiv{\alpha}{\postdist}{\approxdist}  
&\le 
\divbd{\alpha}{\approxdist}{\eta}.
\]
\end{lemma}
Then, combining \cref{thm:Wasserstein-simple,lem:divergence-elbo-cubo-bounds}
yields the desired bounds on the $p$-Wasserstein distance given only the
quantities $\PIC{2p}{\approxdist}$, $\EIC{p}{\approxdist}$, $\cubo{\alpha}{\approxdist}$, and $\elbo{\eta}$,
all of which can be either efficiently estimated or bounded (with high probability);
we address these computational issues in detail in \cref{sec:worflow-detail}.
\begin{theorem} \label{cor:computable-bounds-simple}
For any $p \ge 1$ and any distribution $\eta$, if $\postdist \abscont \approxdist$, then
\[
\pwassSimple{p}{\approxdist}{\postdist} \leq \PIC{2p}{\approxdist}\big[\exp\left\{\divbd{2}{\approxdist}{\eta}\right\} - 1\big]^{\frac{1}{2p}}  \label{eq:computable-PI-bound}
\]
and
\[
\pwassSimple{p}{\approxdist}{\postdist} \le \EIC{p}{\approxdist}\left[\divbd{2}{\approxdist}{\eta}^{\frac{1}{p}} + \{\divbd{2}{\approxdist}{\eta}/2\}^{\frac{1}{2p}}\right]. 
\]
\end{theorem}

\subsection{Importance sampling} \label{sec:importance-sampling}

Before presenting our workflow for validated variational inference, we briefly discuss a final ingredient: importance sampling. 
Standard importance sampling with importance distribution $\approxdist$ operates as follows.
After obtaining samples $\param_{1},\dots,\param_{T} \dist \approxdist$, we can define 
importance weights $w_{t} \defined \postdist^{*}(\param_{t})/\approxdist(\param_{t})$ and self-normalized weights $\tw_{t} \defined w_{t}/\sum_{t=1}^{T}w_{t}$.
Then, the importance sampling estimator for $\int \phi\,\dee\postdist$ is $\sum_{t=1}^{T}\tw_{t}\phi(\param_{t})$.
Importance sampling can decrease the bias at the cost of some additional variance relative to the simple Monte Carlo estimate $T^{-1}\sum_{t=1}^{T}\phi(\param_{t})$.
Recall from \cref{sec:previous-work} that Pareto-smoothed importance sampling (PSIS) can improve upon standard importance sampling by 
significantly reducing variance without much extra bias. 
In additional, PSIS provides a crucial diagnostic quantity, $\hat{k}$.
When $\hat{k} > 0.7$, the importance weights are too high-variance to be reliable, even when using PSIS. 

Our approach to bounding the Wasserstein distance in terms of the $\alpha$-divergence has intriguing connections to 
the theory of importance sampling.
As pointed out by \citet{Dieng:2017}, minimizing the 2-divergence is equivalent to minimizing the variance of the (normalized) importance weight 
$\postdist(\param_{t})/\approxdist(\param_{t})$, which is equal to $\exp\{\alphaDiv{2}{\targetdist}{\approxdist}\} - 1$. 
Moreover, the estimation error of importance sampling can be bounded as a function of $\kl{\targetdist}{\approxdist}$~\citep{Chatterjee:2018}, which is upper bounded by 
$\alphaDiv{2}{\targetdist}{\approxdist}$.
Thus, minimizing the 2-divergence simultaneously leads to better importance distributions and smaller Wasserstein error -- as long as the 
moments of the variational approximation do not increase disproportionately to the 2-divergence decrease.
In practice such pathological behavior appears to be unusual; see \cref{sec:experiments} and \citet[\S3]{Dieng:2017}.

\subsection{A workflow for black box variational inference} \label{sec:worflow-detail}

The usual approach to black box variational inference is 
(1) to choose $\discrepancy{\postdist}{\vardist} = \kl{\vardist}{\postdist}$ (i.e., to maximize $\elbo{\vardist}$), and
(2) to use (products of) Gaussians  as the variational family $\varfamily$ \citep{Ranganath:2014,Kucukelbir:2015,Salvatier:2016,Carpenter:2017}. 
Based on \cref{cor:computable-bounds-simple} and our discussion in \cref{sec:importance-sampling},
we suggest a number of deviations from the typical variational inference procedure,
including integrating checks based on our novel bounds. 
We first provide an outline of our default workflow recommendation, 
then discuss each step in detail -- along with some potential refinements.
We show the workflow in action in \cref{sec:experiments}.
We write $\varfamily_{h}^{\distT}$ to denote the mean-field variational family consisting of product of $t$-distributions with $h$ degrees of freedom.

\IncMargin{1em}
\begin{algorithm}[h]
\nl Set $\varfamily$ to be $\varfamily_{40}^{\distT}$\;
\nl Find $\approxdist \in \varfamily$ that minimizes \cubo{2}{\vardist} \label{step:CHIVI} \;
\nl\If{$\hat{k} > 0.7$}{
	Refine choice of $\varfamily$ or reparameterize the model\; %
	Return to step \ref{step:CHIVI}}
\nl Find $\eta \in \varfamily$ that maximizes $\elbo{\vardist}$\;
\nl Estimate \elbo{\eta}\ and \cubo{2}{\approxdist} via Monte Carlo\;
\nl Use \cref{lem:divergence-elbo-cubo-bounds} to compute bound $\bar{\delta}_{2} \ge \alphaDiv{2}{\postdist}{\approxdist}$\;
\nl Use \cref{cor:computable-bounds-simple} to compute bound $\bar{w}_{2} \ge \pwassSimple{2}{\postdist}{\approxdist}$\;
\nl\uIf{$\bar{\delta}_{2}$ and $\bar{w}_{2}$ are large}{
	Refine choice of $\varfamily$ or reparameterize the model\;
	Return to step \ref{step:CHIVI}} %
\uElseIf{$\bar{\delta}_{2}$ and $\bar{w}_{2}$ are very small}{
	approximate $\postdist$ with $\approxdist$}
\Else({when $\bar{\delta}_{2}$ and $\bar{w}_{2}$ are moderately small}){
	Use %
	PSIS to refine the posterior expectations produced by $\approxdist$}
\caption{Validated variational workflow}
\end{algorithm}
\DecMargin{1em}

For step 1, we choose a heavy-tailed variational family to ensure that the 2-divergence (and hence $\cubo{2}{\vardist}$) is finite
(that is, such that $\alphaDiv{2}{\targetdist}{\vardist} < \infty$ for all $\vardist \in \varfamily$).
The choice of 40 degrees of freedom is somewhat arbitrary. Slightly different choices should produce similar results. 
It is also possible to select a different variational family specific to the problem at hand as long as the 2-divergence is guaranteed to be finite.
For step 2, we minimize the CUBO to obtain as tight a bound as possible when we apply \cref{cor:computable-bounds-simple}
(though note that usually the CUBO objective -- like the negative ELBO -- is non-convex, so we may not be able to find the global minimum).
Toward the same end, in step 4 we separately find the distribution $\eta$ that results in largest ELBO. 
However, before going to the effort of finding $\eta$, in step 3 we check that $\hat{k} \le 0.7$, since otherwise 
our estimate of \cubo{2}{\approxdist} is not reliable and thus we should not trust any bounds on the 2-divergence or Wasserstein distance
computed using \cref{lem:divergence-elbo-cubo-bounds,cor:computable-bounds-simple}. 
How precisely to refine the choice of $\varfamily$ or reparameterize the model is problem-dependent. 
One possibility is to use multivariate $t$-distributions with $h$ degrees of freedom for $\varfamily$; unlike $\varfamily_{h}^{\distT}$, the multivariate versions
can capture correlations in the posterior. 

For step 5, we can use simple Monte Carlo to compute high-accuracy estimates for \elbo{\eta} and \cubo{2}{\approxdist}:
\[
\cuboEst{2}{\approxdist} &\!\defined\! \textstyle\frac{1}{2}\log \big[\frac{1}{T}\sum_{t=1}^{T} \big\{\der{\postdist^{*}}{\approxdist}(\theta^{\approxdist}_{t})\big\}^2\big], \!\!\!\!\!&
	(\theta^{\approxdist}_{t})_{t=1}^{T} \distiid \approxdist \\
\elboEst{\eta} &\!\defined\! \textstyle\frac{1}{T}\sum_{t=1}^{T}\log \der{\postdist^{*}}{\eta}(\theta^{\eta}_{t}), \!\!\!&
	(\theta^{\eta}_{t})_{t=1}^{T} \distiid \eta.
\]
Ensuring the accuracy of $\cuboEst{2}{\approxdist}$ and $\elboEst{\eta}$ reduces to the well-studied problem of estimating the accuracy of
a simple Monte Carlo approximation~\citep[e.g.,][]{Koehler:2009}. 
We can also convert these estimates into high-probability upper bounds using standard concentration inequalities \citep{Boucheron:2013}.
For step 6, we use \cref{lem:divergence-elbo-cubo-bounds} to obtain the estimated 2-divergence bound
\[
\bar{\delta}_{2} \defined \divbdEst{2}{\approxdist}{\eta} \defined 2\big\{\cuboEst{2}{\approxdist} - \elboEst{\eta}\big\}.
\]
For step 7, to compute Wasserstein bounds using \cref{cor:computable-bounds-simple}, we can bound $\PIC{2p}{\approxdist}$ using the central moments of 
the distribution: if $\approxdist = \prod_{i=1}^{d}\distT_{h}(\mu_{i}, \sigma_{i})$ and $C_{h} \defined h/(h-2)$, then 
\[
\PIC{2}{\approxdist} &\le \textstyle 2C_{h} \sum_{i=1}^{d}\sigma_{i}^{2} \\
\PIC{2}{\approxdist} &\le \textstyle 2C_{h}^{2}\left\{ \frac{2(h-1)}{h-4}\sum_{i=1}^{d}\sigma_{i}^{2} + (\sum_{i=1}^{d}\sigma_{i}^{2})^{2}\right\}.
\]
Since $\EICest{p}{\approxdist} = \infty$ for $t$-distribution variational families, we cannot use the second bound from \cref{cor:computable-bounds-simple}.
For variational families without analytically computable moments, we can bound the moment constants $\PIC{p}{\approxdist}$ and $\EIC{p}{\approxdist}$ by fixing any
 $\theta_0, \epsilon$ and sampling from $\approxdist$.
We can intuitively think of $\theta_{0}$ as the ``center'' of the distribution, so a natural choice is setting it equal to the mean of $\approxdist$.

For step 8, what qualifies as a moderately or very small $\bar{w}_{2}$ value will depend on the desired accuracy and natural scale of the problem.
$\bar{\delta}_{2}$ has a more universal scale; in particular, $\bar{\delta}_{2} < 4.6$ could be treated as moderately small since
the variance of the importance weights is $\exp\{\alphaDiv{2}{\approxdist}{\postdist})\} - 1 < 100$, so PSIS with a reasonable number of samples should be effective;
for some $\delta_{*} \ll 1$ (for example, $ \delta_{*} = 0.01$), $\bar{\delta}_{2} < \delta_{*}$ could be treated as very small, since the term multiplying $\PIC{p}{\vardist}$ 
in \cref{thm:Wasserstein-simple,cor:computable-bounds-simple} will be less than $\delta_{*}^{1/p}$.

%% file: experiments.tex
\section{Two case studies} \label{sec:experiments}

Next we demonstrate our variational inference workflow and the usefulness
of our %
bounds through two case studies.

\subsection{Case study \#1: the eight schools model}

We apply our variational workflow to approximate the posterior for the eight schools data and
model~\citep[Sec.~5.5]{Gelman:2013}, a canonical
example of a Bayesian hierarchical analysis. \citet{Yao:2018} previously considered this model
in the setting of evaluating variational inference.
In the eight schools data, we have observations corresponding to the mean $y_n$ and 
standard deviation $\sigma_n$ of a treatment effect at each of eight schools, indexed by $n\in\{1,\dots,8\}$.  The 
goal is to estimate the overall treatment effect $\mu$, the standard deviation $\tau$ of school-level
treatment effects, and the true school-level treatment effects $\theta_{n}$.
There are two standard ways to parameterize the model.
The \emph{centered parameterization} is 
\[
y_{n} \given \theta_{n} &\dist \distNorm(\theta_{n}, \sigma_{n}),  & \!\!\!
\theta_{n} \given \mu, \tau & \dist \distNorm(\mu, \tau), \label{eq:8schools}  \\
\mu &\dist \distNorm(0, 5), & \tau &\dist  \distNamed{half{\mhyphen}Cauchy}(0,5).
\]
The \emph{non-centered parameterization} decouples $\theta$ and $\tau$
through the transformation $\ttheta_{n} = (\theta_{n} - \mu_{n})/\tau$ and replaces
\cref{eq:8schools} with
\[
y_{n} \given \ttheta_{n} &\dist \distNorm(\mu + \tau\ttheta_{n}), & 
\ttheta_{n} &\dist \distNorm(0,1).
\]
The standard deviation of the $y_{n}$ is $9.8$ and the median $\sigma_{n}$ value is 11, 
which suggests the overall scale of the problem is roughly 10.

\textbf{Diagnosing a poorly parameterized model.}
We begin by considering the centered parameterization. 
Following steps 1 and 2, we use $\varfamily = \varfamily_{40}^{\distT}$
and minimize the CUBO using CHIVI \citep{Dieng:2017} to obtain $\approxdist$.
The results appear in the first column of \cref{tbl:8-schools}. 
Since $\hat{k} > 0.7$, according to step 3  we should either reparameterize the model or choose a different $\varfamily$.
If we compute $\bar\delta_{2}$ and $\bar w_{2}$, then we reach the same conclusion.
Thus, these diagnostics correctly determine that $\approxdist$ is a poor approximation to
the posterior and that PSIS does not provide an improvement.

\begin{figure}[tbp]
\begin{center}
\begin{subfigure}[t]{.49\textwidth}
\centering
\includegraphics[width=\textwidth]{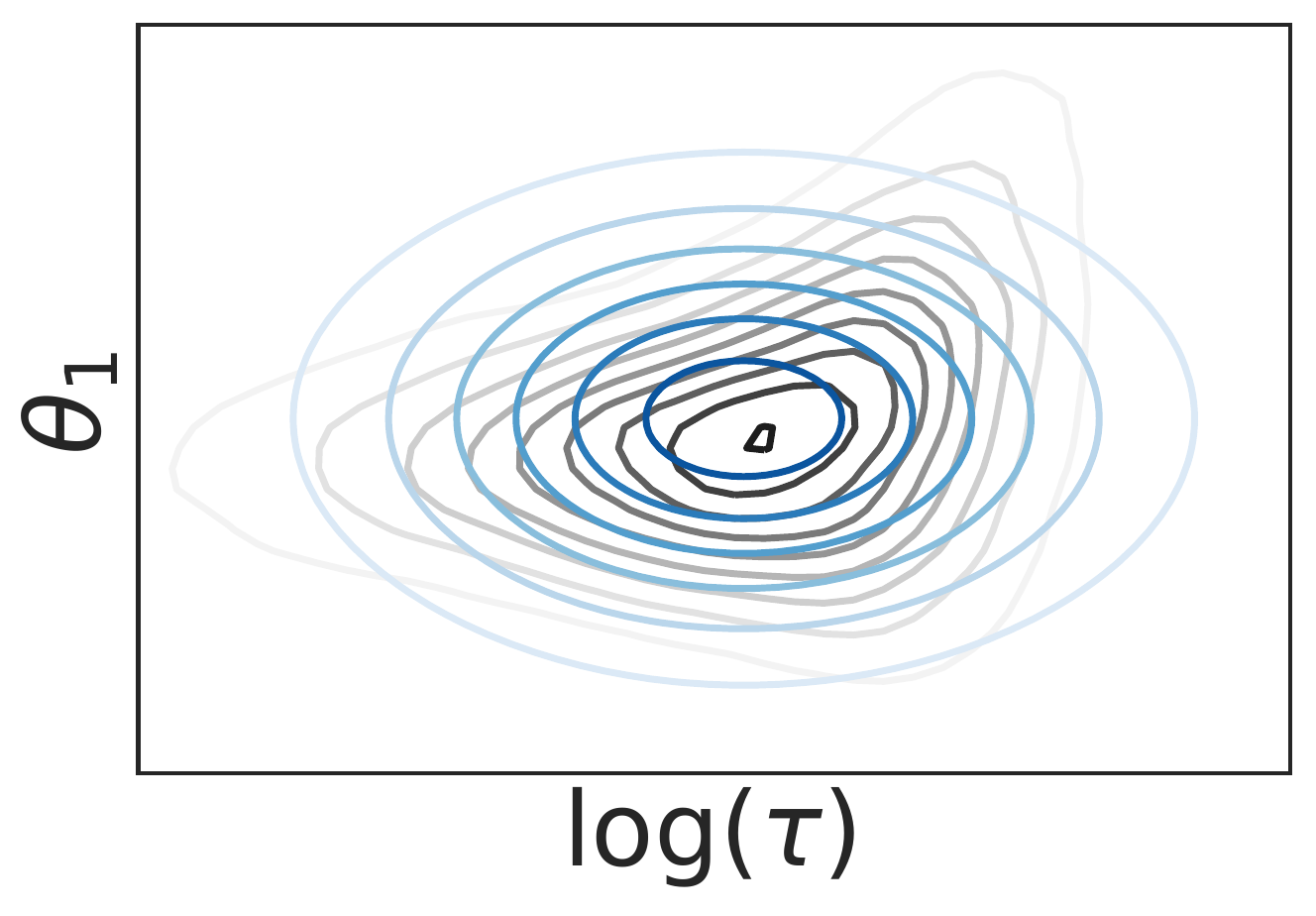}
\caption{centered} %
\end{subfigure}  
\begin{subfigure}[t]{.49\textwidth}
\centering
\includegraphics[width=\textwidth]{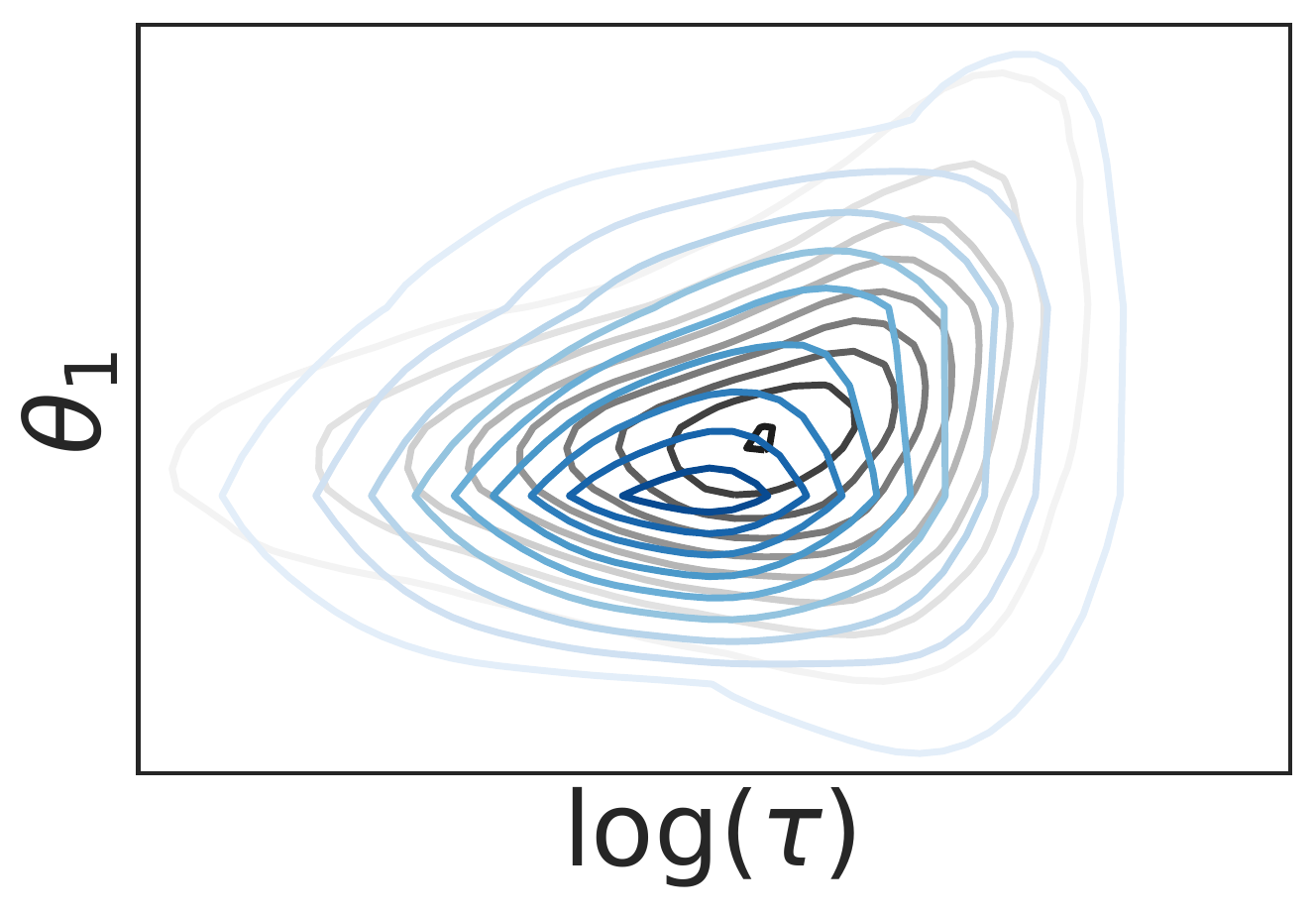}
\caption{non-centered} %
\end{subfigure} 
\caption{Approximate posteriors for CHIVI variational approximations (blue) and 
for HMC (black; ground truth).}
\label{fig:8-schools-posterior-approximations}
\end{center}
\end{figure}

\begin{table}[t]
\begin{tabular}{@{}lclcc@{}}
\toprule
                           & centered &  & \multicolumn{2}{c}{non-centered} \\
                           & df = 40  &  & df = 40         & df = 8         \\ \midrule
$D_2$ bound $\bar\delta_2$ & 14       &  & 1.6             & 3.8            \\
$\hat{k}$                  & 0.88     &  & 0.55            & 0.40           \\
$W_2$ bound $\bar w_2$     & 983      &  & 15              & 29             \\
mean error                 & 0.10     &  & 0.14            & 0.18           \\
\quad with PSIS            & 0.40     &  & 0.04            & 0.03           \\
std. dev. error            & 1.1      &  & 0.03            & 0.25           \\
\quad with PSIS            & 1.2      &  & 0.03            & 0.02           \\
covariance error           & 8.4      &  & 0.95            & 1.3            \\
\quad with PSIS            & 5.4      &  & 0.45            & 0.32           \\ \bottomrule
\end{tabular}
 \caption{Results for eight schools model for the parameter vector $(\mu, \log \tau, \theta_{1}, \dots, \theta_{8})$. 
The mean and standard deviation errors are defined as, respectively, $\twonorm{\mean{\postdist} - \mean{\approxdist}}$
and $\twonorm{\sigma_{\postdist} - \sigma_{\approxdist}}$.
The covariance error is defined as $\twonorm{\Sigma_{\postdist} - \Sigma_{\approxdist}}^{1/2}$.
We use the square root for the covariance error in order to place it on the same scale 
as the mean error, standard deviation error, and the 2-Wasserstein bound. 
To provide a sense of the overall scale, we note that $\twonorm{\Sigma_{\postdist}}^{1/2} = 9.7$.}
\label{tbl:8-schools}%
\end{table}

\begin{table}[]
\begin{tabular}{@{}lccc@{}}
\toprule
                 & \begin{tabular}[c]{@{}c@{}}mean-field \\ KLVI\end{tabular} & \begin{tabular}[c]{@{}c@{}}mean-field\\ CHIVI\end{tabular} & \begin{tabular}[c]{@{}c@{}}full-rank\\ KLVI\end{tabular} \\ \midrule
$D_2$ bound      & 8.7                                                        & 4.9                                                        & $6 \times 10^{-3}$                                         \\
$\hat{k}$        & 0.92                                                       & 0.34                                                       & -0.93                                                    \\
$W_2$ bound      & 4.4                                                        & 8.4                                                        & 0.39                                                     \\
mean error       & 0.01                                                       & $<0.01$                                                    & 0.01                                                     \\
\quad with PSIS  & 0.06                                                       & 0.01                                                       & 0.01                                                     \\
std. dev. error  & 0.73                                                       & 0.09                                                       & $<0.01$                                                  \\
\quad with PSIS  & 0.49                                                       & $<0.01$                                                    & $<0.01$                                                  \\
covariance error & 0.92                                                       & 0.72                                                       & $<.1$                                                    \\
\quad with PSIS  & 0.82                                                       & 0.11                                                       & $<.1$                                                    \\ \bottomrule
\end{tabular}
\caption{Results for robust regression. See \cref{tbl:8-schools} for further explanation. 
To provide a sense of the overall scale, we note that $\twonorm{\Sigma_{\postdist}}^{1/2} = 0.93$.
}
\label{tbl:robust-regression}
\end{table}

\textbf{Correctly validating an improved parameterization.}
\cref{fig:8-schools-posterior-approximations}a compares approximate posteriors
from CHIVI and Hamiltonian Monte Carlo \citep[HMC;][]{neal2011mcmc} --
namely the dynamic HMC implementation in Stan \citep{hoffman2014no,carpenter2017stan}. 
The HMC samples serve as ground truth.
This comparison illustrates why the centered parameterization is not
conducive to variational inference when using a mean-field variational family: the conditional variance of any $\theta_{n}$ is strongly
dependent on $\tau$. 
We can remedy this issue by using the non-centered parameterization.
Repeating steps 1 and 2, we now find that $\hat{k} = 0.55 \le 0.7$, suggesting
the variational approximation is at least acceptable as an importance distribution (step 3). 
However, $\bar\delta_{2}$ and $\bar w_{2}$ remain at best moderately small, 
so the variational approximation should not be used directly (steps 4--8). 
These diagnostic results are confirmed graphically in \cref{fig:8-schools-posterior-approximations}b
and quantitatively in the second column of \cref{tbl:8-schools}.
Furthermore, applying PSIS does reduce approximation error, as expected.

\begin{figure*}[tbp]
\begin{center}
\begin{subfigure}[t]{.32\textwidth}
\centering
\includegraphics[width=\textwidth]{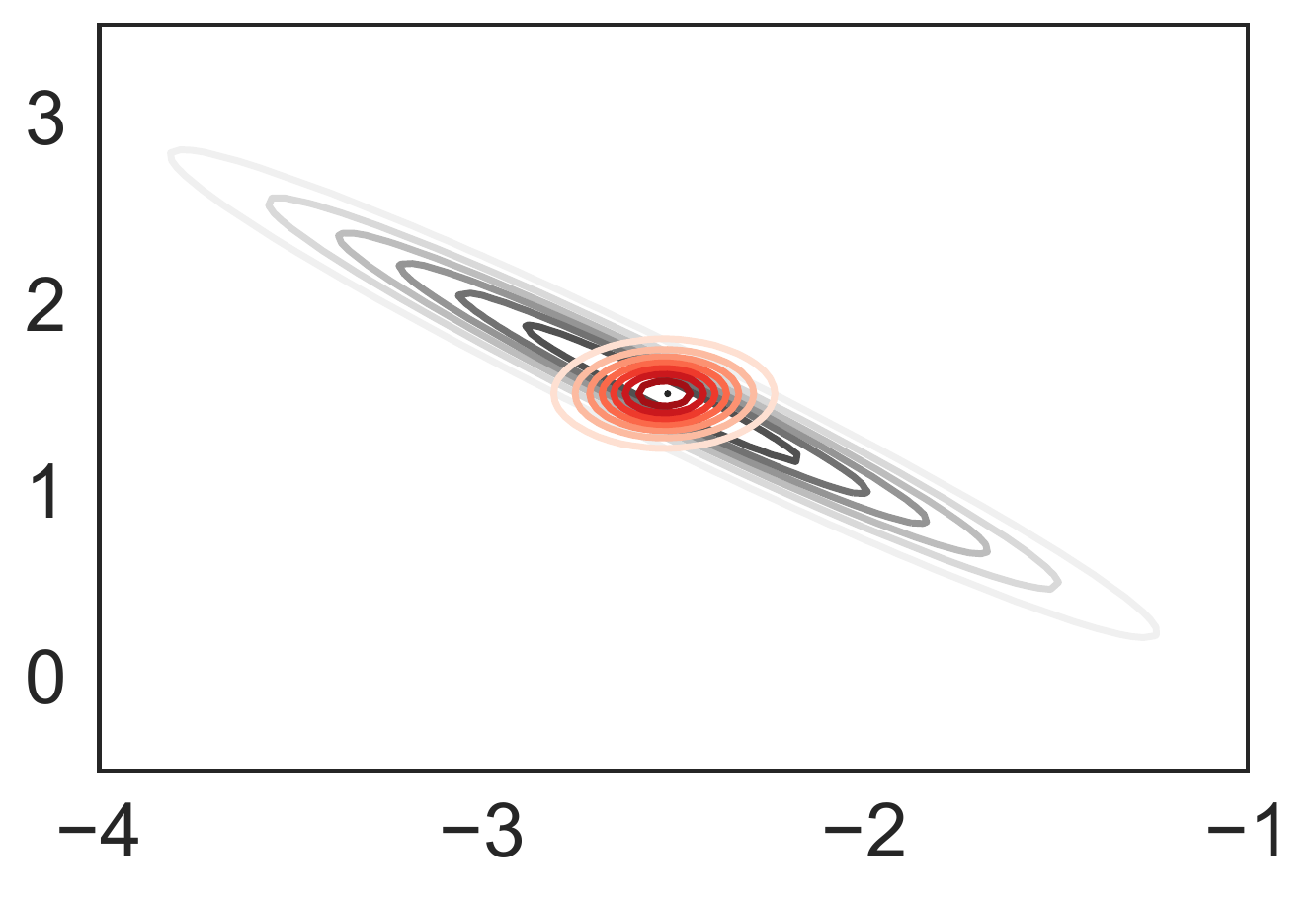}
\caption{mean-field KLVI}
\end{subfigure}
\begin{subfigure}[t]{.32\textwidth}
\centering
\includegraphics[width=\textwidth]{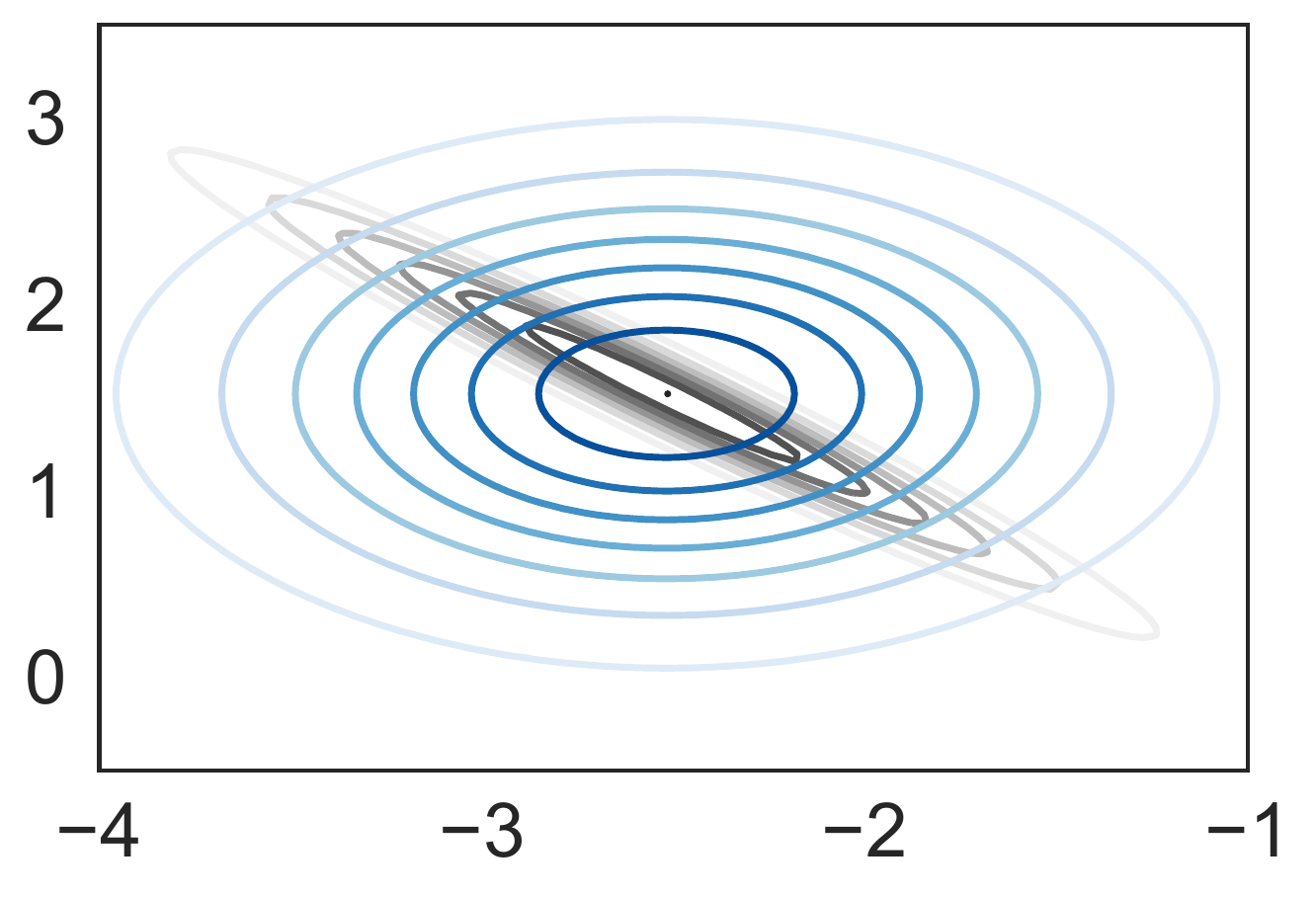}
\caption{mean-field CHIVI}
\end{subfigure} 
\begin{subfigure}[t]{.32\textwidth}
\centering
\includegraphics[width=\textwidth]{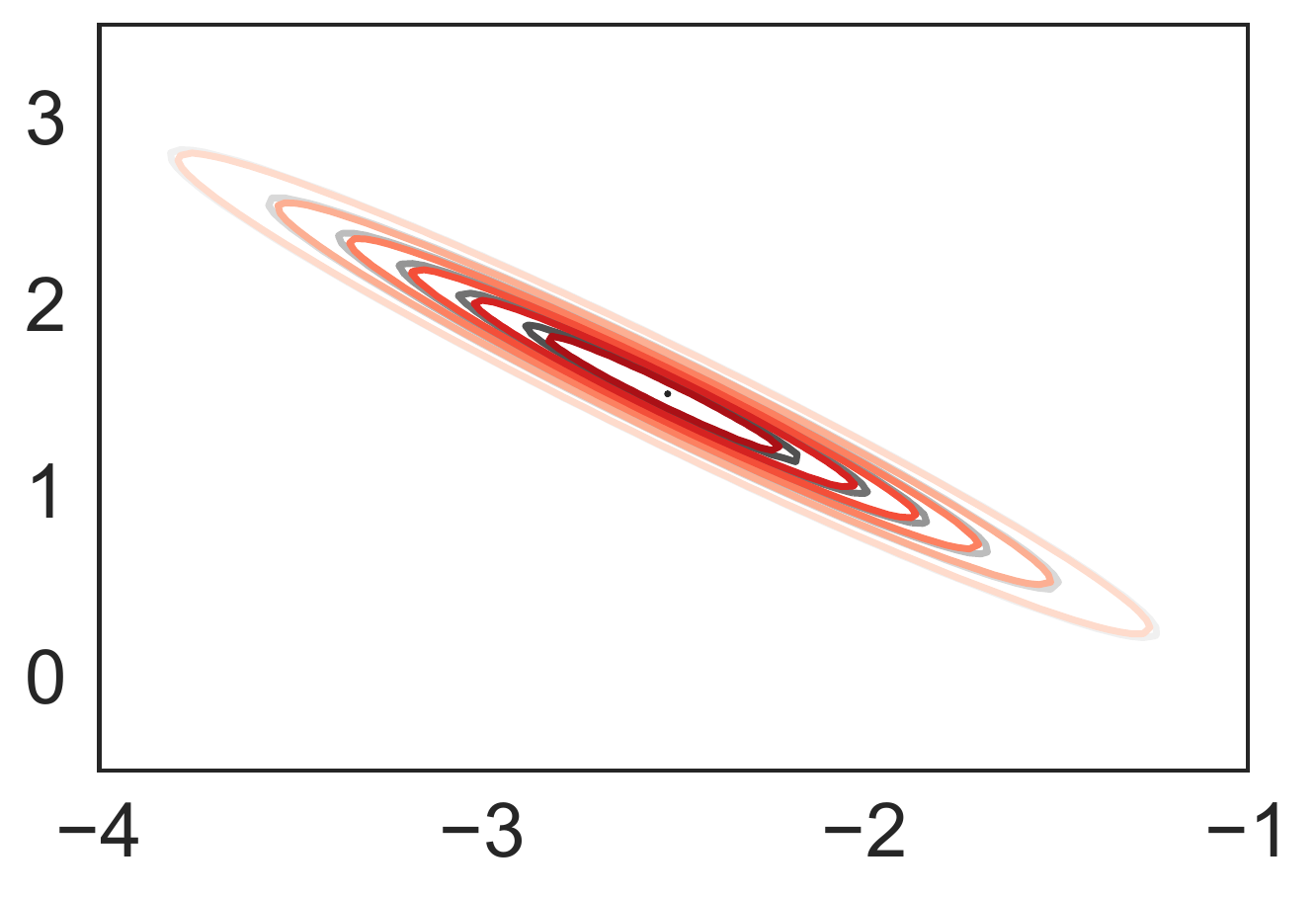}
\caption{full-rank KLVI}
\end{subfigure}
\caption{For robust regression, approximate posteriors for variational approximations (KLVI red, CHIVI blue) and 
for the exact posterior (black).}
\label{fig:robust-regression-posterior-approximations}
\end{center}
\end{figure*}

\textbf{Limitations of $\hat{k}$.}
So far $\hat{k}$, $\bar\delta_{2}$, and $\bar w_{2}$ have all provided similar diagnostic information.
To see how they can diverge, we repeat our workflow, but this time we use $\varfamily = \varfamily_{8}^{\distT}$.
The results appear in the final column of \cref{tbl:8-schools}. 
Variational approximations with heavier tails should decrease $k$ and $\hat{k}$ since the
importance weights will have more finite moments.  
While $\approxdist \in \varfamily_{8}^{\distT}$ has a small $\hat{k}$ value (0.40),
its accuracy is worse than the non-centered $\approxdist \in \varfamily_{40}^{\distT}$ ($\hat{k} = 0.55$). 
The lower quality of $\approxdist \in \varfamily_{8}^{\distT}$ is, however, accurately reflected
in the larger $\bar\delta_{2}$ and $\bar w_{2}$ values. 
The accuracy of PSIS using $\approxdist \in \varfamily_{8}^{\distT}$, however, is better than that of PSIS using 
the non-centered $\approxdist \in \varfamily_{40}^{\distT}$.
In sum, while $\hat{k}$ does provide a useful diagnostic for when $\approxdist$
will serve as a good importance distribution, it does not provide a reliable
heuristic for the accuracy of $\approxdist$ as an approximation to $\postdist$.
Hence, our workflow uses $\hat{k}$ only to validate the use of importance sampling, but not the quality of $\approxdist$.

\subsection{Case study \#2: robust regression} \label{sec:robust-regression}

A second example of our variational inference workflow confirms our findings in the eight schools example
and further clarifies the differences between $\hat{k}$ and our bounds.
Specifically, we consider the posterior of the toy robust regression model
with coefficients $\theta \in \reals^{d}$ and observed data $(x_{1}, y_{1}), \dots, (x_{N}, y_{N}) \in \reals^{d+1}$:
\[
\theta_{i} &\dist \distNorm(0, 10), &
y_{n} \given x_{n}, \theta &\dist \distT_{40}(\beta^{\top}x_{n}, 1). \label{eq:robust-reg}
\]
We take $d=2$ and $N=25$. We generate data according to \cref{eq:robust-reg} with $\beta = (-2,1)$ and each $x_{n}$
Gaussian-distributed, with $\mathrm{var}(x_{ni}) = 1$ and $\cov(x_{n1}, x_{n2}) = 0.75$.

\textbf{A poor quality approximation.} 
For illustrative purposes, first we approximate the posterior with standard black-box variational inference (i.e., by maximizing the ELBO).
We refer to this method as KLVI. 
As before, we use $\varfamily = \varfamily_{40}^{\distT}$. 
The results appear in the first column of \cref{tbl:robust-regression}.
The variational approximation is poor due to the strong posterior correlation between $\theta_{1}$ and $\theta_{2}$ (\cref{fig:robust-regression-posterior-approximations}a),
and PSIS is unable to correct for the underdispersion of the posterior approximation. 
The large values of $\bar d_{2}$, $\bar w_{2}$, and $\hat{k}$ accurately reflect these findings.

\textbf{$\hat{k}$ does not detect a poor quality CHIVI approximation.}
Next, we instead use CHIVI with $\varfamily_{40}^{\distT}$. 
The results appear in the second column of \cref{tbl:robust-regression}.
The variational approximation is in some sense better because it no longer underestimates the marginal variances (\cref{fig:robust-regression-posterior-approximations}b),
which is reflected in the smaller standard deviation error.
However, since the mean-field family cannot capture the posterior correlation structure, the covariance error remains large.
The poor covariance approximation is reflected in large $\bar d_{2}$ and $\bar w_{2}$ values;
however, it is not reflected in the small $\hat{k}$ value (0.34).

\textbf{The complementary roles of $\bar \delta_{2}$ and $\bar w_{2}$ versus $\hat{k}$.}
We have just seen that the small $\hat{k}$ value when using CHIVI with a mean-field family does not reflect the quality of the variational approximation.
However, the PSIS errors are small, so $\hat{k}$ \emph{does} accurately capture the fact that the CHIVI approximation is a good importance sampling distribution. 
Thus, $\hat{k}$ provide complementary information to $\bar \delta_{2}$ and $\bar w_{2}$.
This complementarity is further illustrated when using KLVI with a full-rank variational family 
(\cref{fig:robust-regression-posterior-approximations}c and the final column of \cref{tbl:robust-regression}).
Now the posterior approximation is very accurate, which is reflected in the very small $\bar d_{2}$ value and fairly small $\bar w_{2}$ value. 
Step 8 of our workflow suggests not using PSIS when $\bar d_{2}$ and $\bar w_{2}$ are small since it will be difficult to improve the posterior
approximation accuracy. 
Applying PSIS confirms that importance sampling is not necessary: although the $\hat{k}$ value is extremely small (in fact, negative),
PSIS and $\approxdist$ provide nearly identical accuracy.

\section{Conclusion}
 
In conclusion, as we have shown through both theory and experiment, our
workflow for validated variational inference potentially provides a framework for making
variational methods more competitive with Markov chain Monte Carlo. We end by
noting that our work complements recent proposals for making variational
approximations arbitrarily
accurate~\citep{Guo16,Wang16,Miller17,Locatello18,Locatello18b,Campbell19}
since our bounds can provide a stopping criteria for when a variational
approximation no longer needs to be improved.

%% file: discussion.tex
\subsubsection*{Acknowledgements}

The authors thank Sushrutha Reddy for pointing out some improvements to our Wasserstein bounds on the standard deviation and variance, 
and also Daniel Simpson, Lester Mackey, Arthur Gretton, and Pierre Jacob for valuable discussions and many useful references. 
M.\ Kasprzak was supported in part by an EPSRC studentship and FNR grant FoRGES (R-AGR-3376-10).
T.\ Campbell was supported by a National Sciences and Engineering Research Council of Canada (NSERC)
Discovery Grant and Discovery Launch Supplement.
T.\ Broderick was supported in part by an NSF CAREER Award, an ARO YIP Award, the Office of Naval Research, a Sloan Research Fellowship, the CSAIL-MSR Trustworthy AI Initiative, and DARPA.

%% file: appendix.tex
\onecolumn
\begin{adjustwidth}{.5in}{.5in}
\appendix

\begin{figure*}[tbp]
\begin{center}
\begin{subfigure}[b]{.28\textwidth}
\centering
\includegraphics[width=\textwidth]{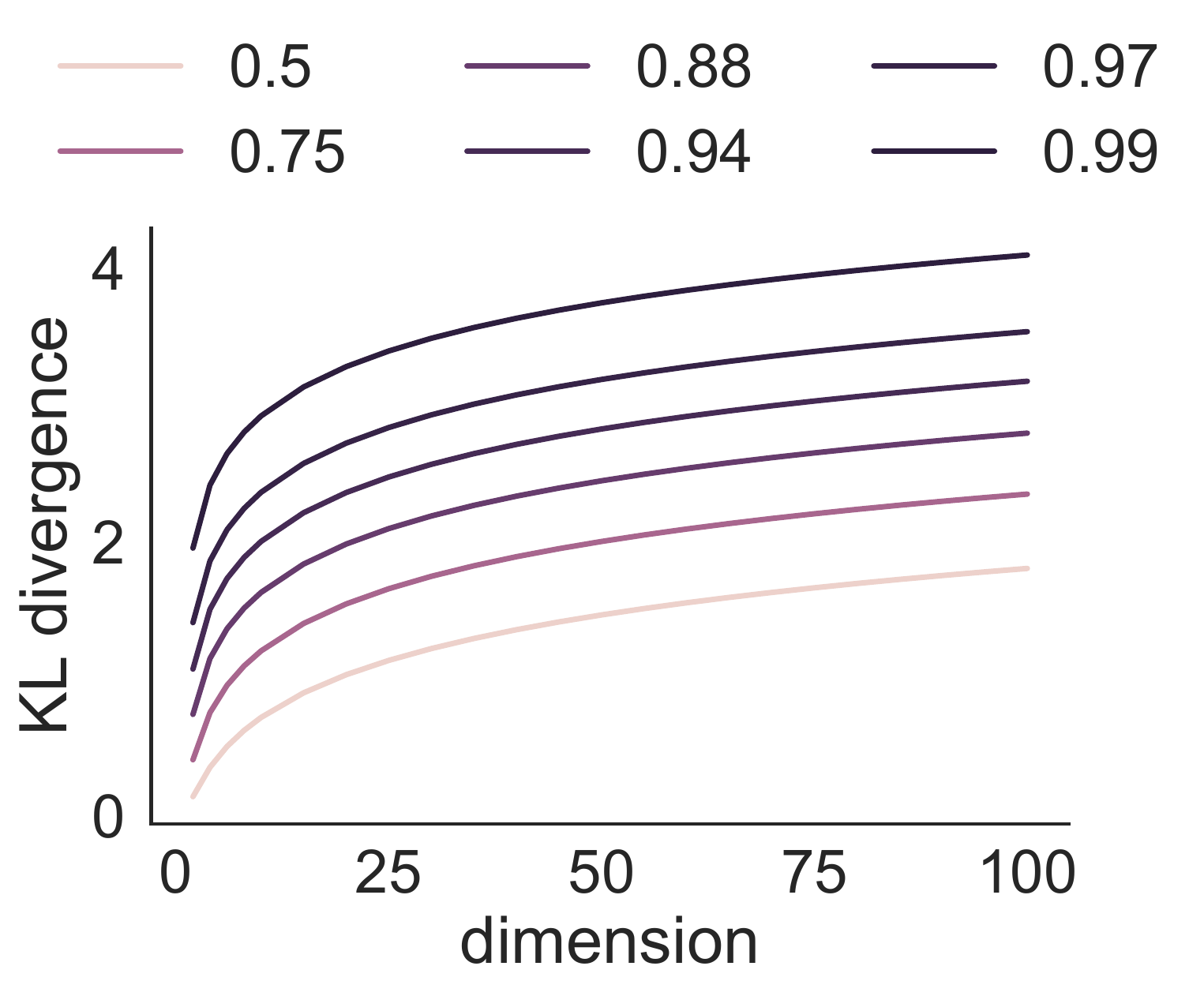}
\vspace{1.5em}
\caption{}
\end{subfigure}
\begin{subfigure}[b]{.71\textwidth}
\centering
\includegraphics[width=.32\textwidth]{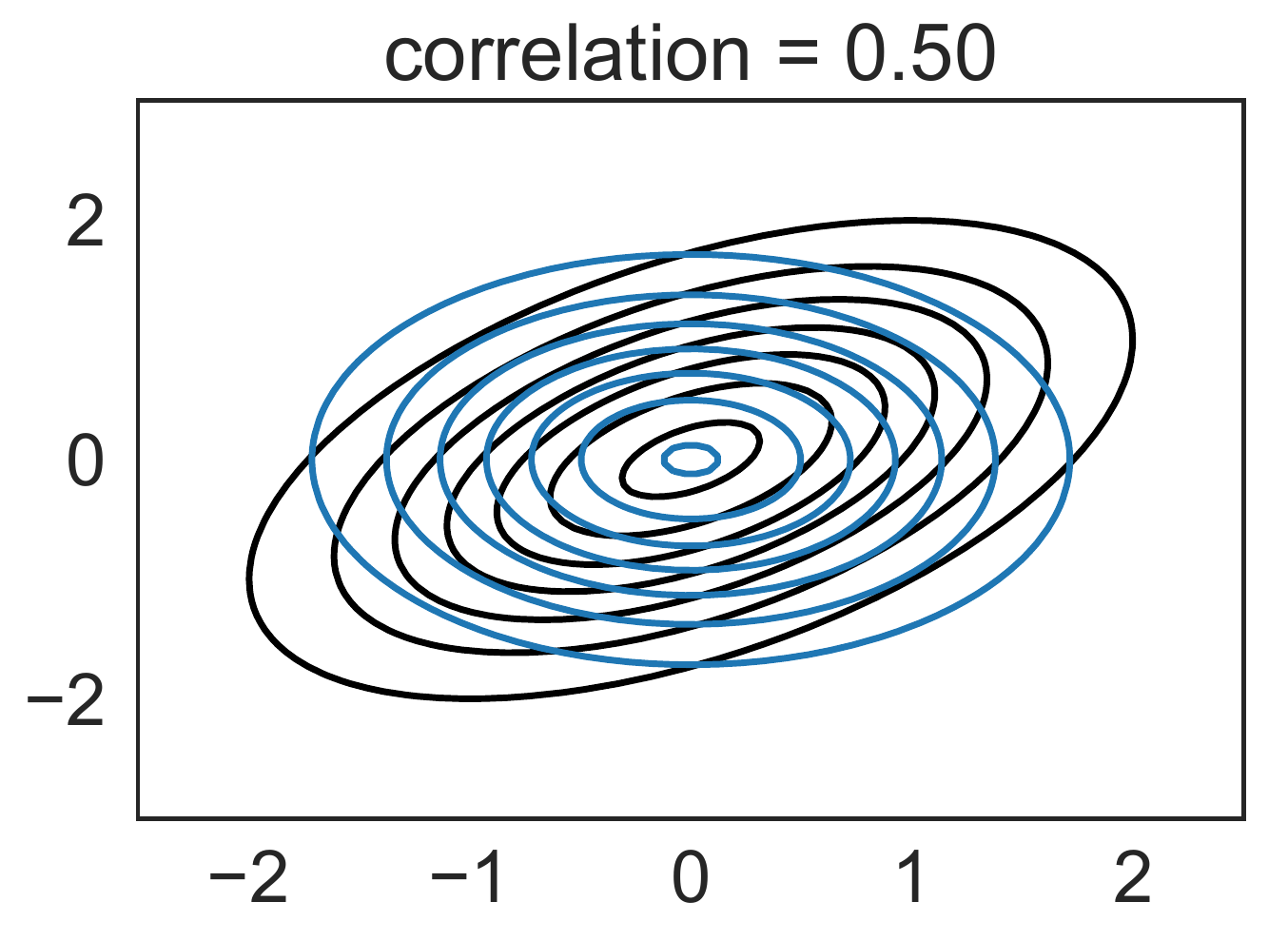}
\includegraphics[width=.32\textwidth]{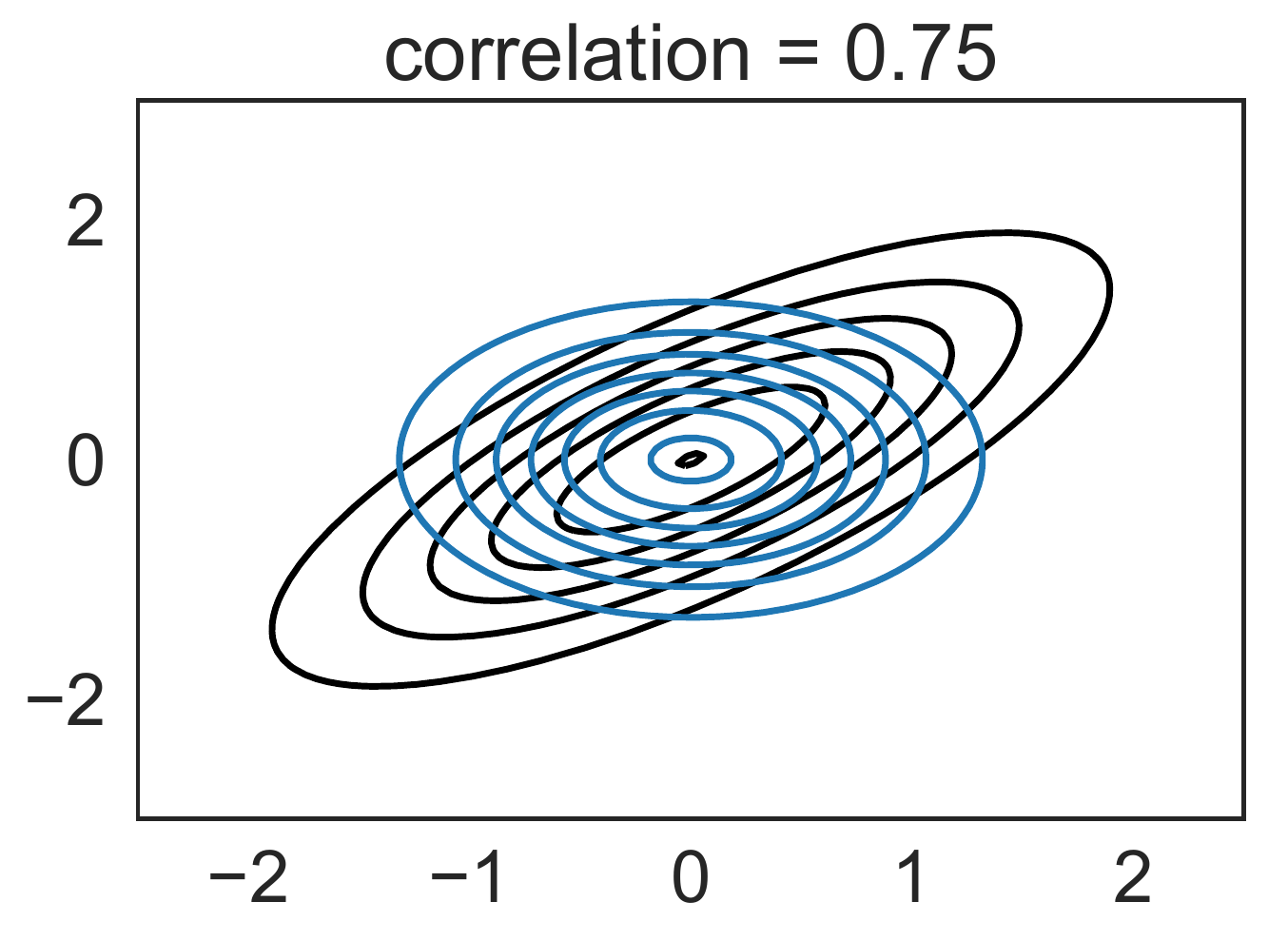}
\includegraphics[width=.32\textwidth]{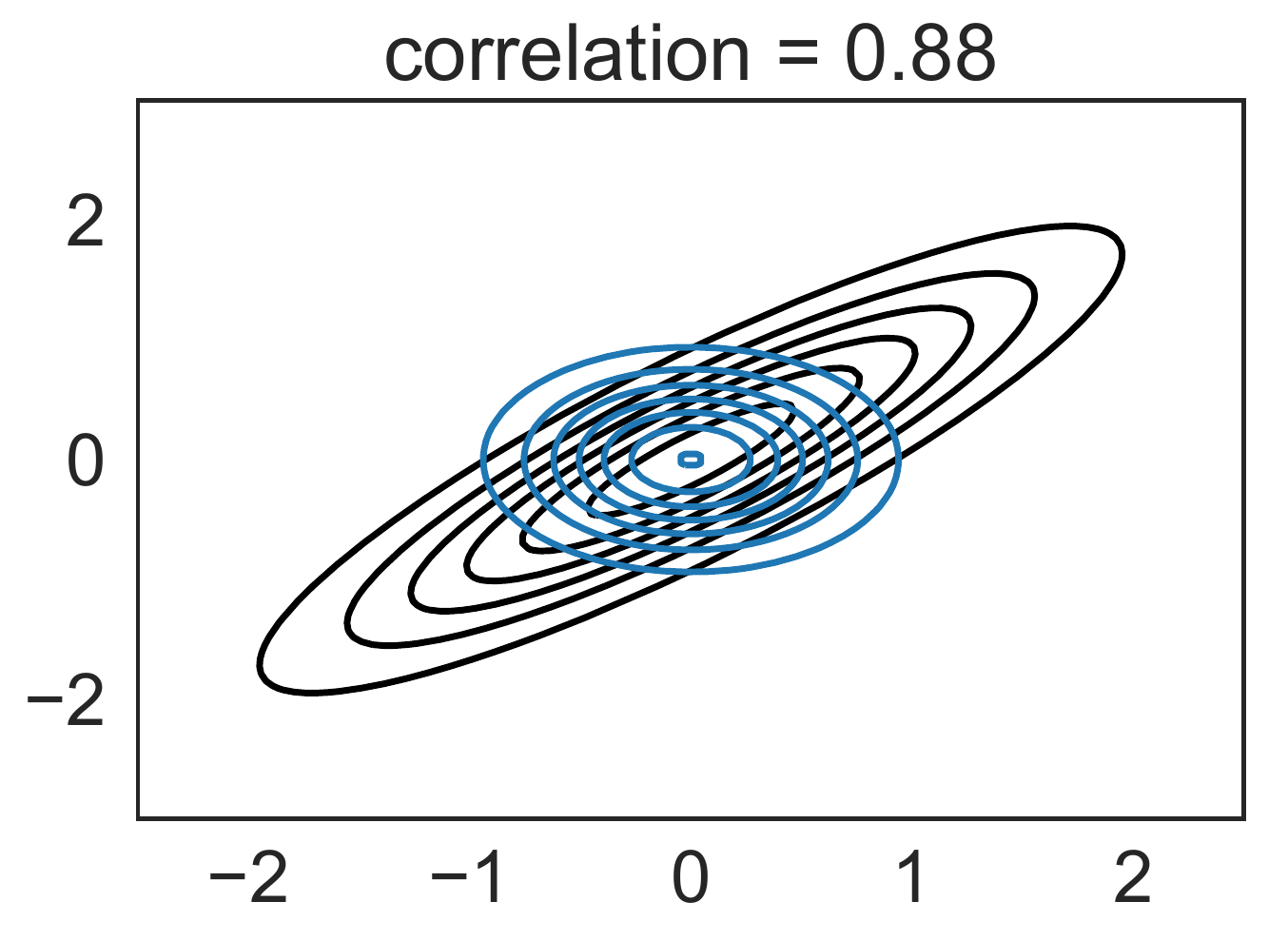} \\
\includegraphics[width=.32\textwidth]{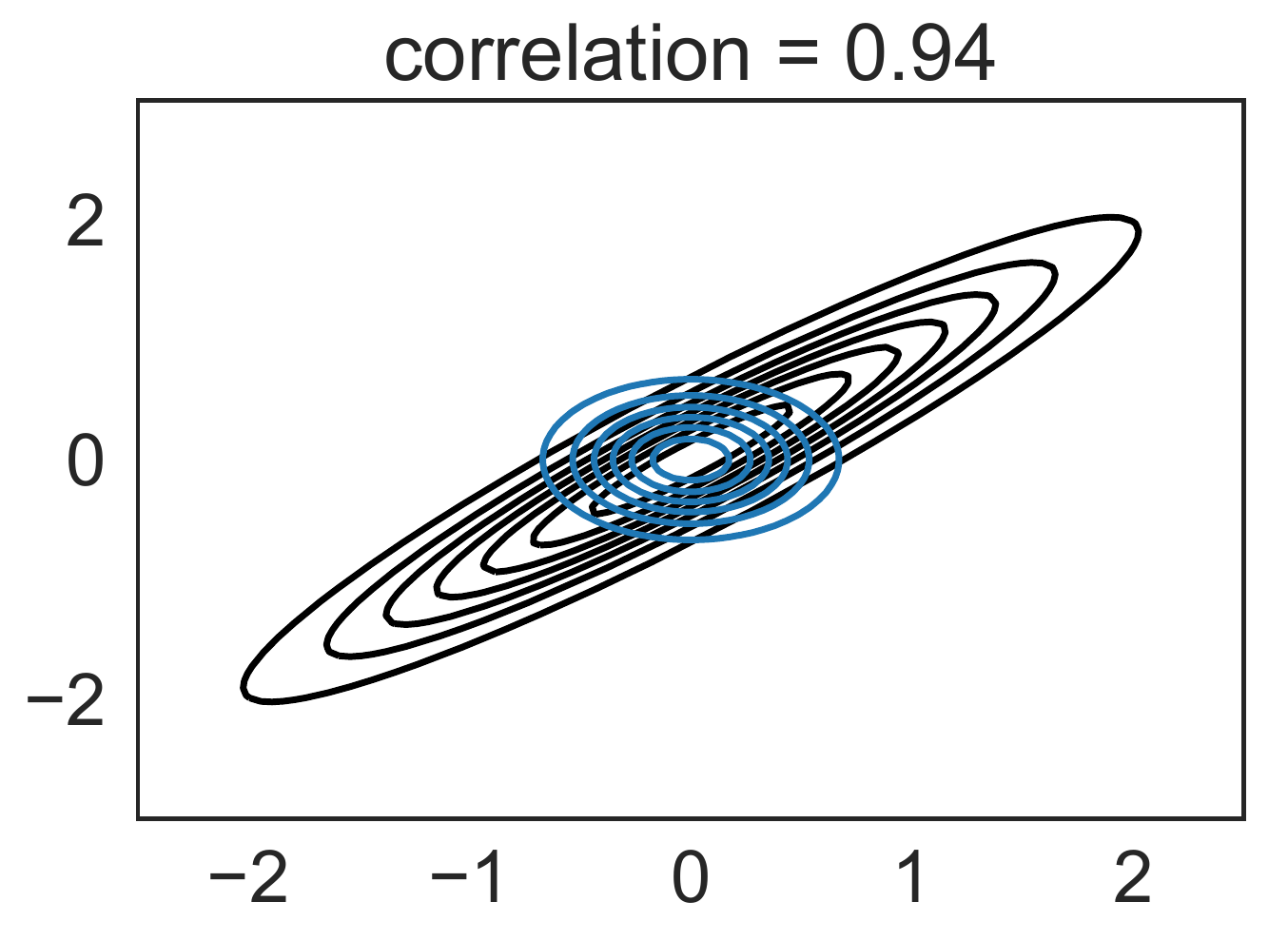}
\includegraphics[width=.32\textwidth]{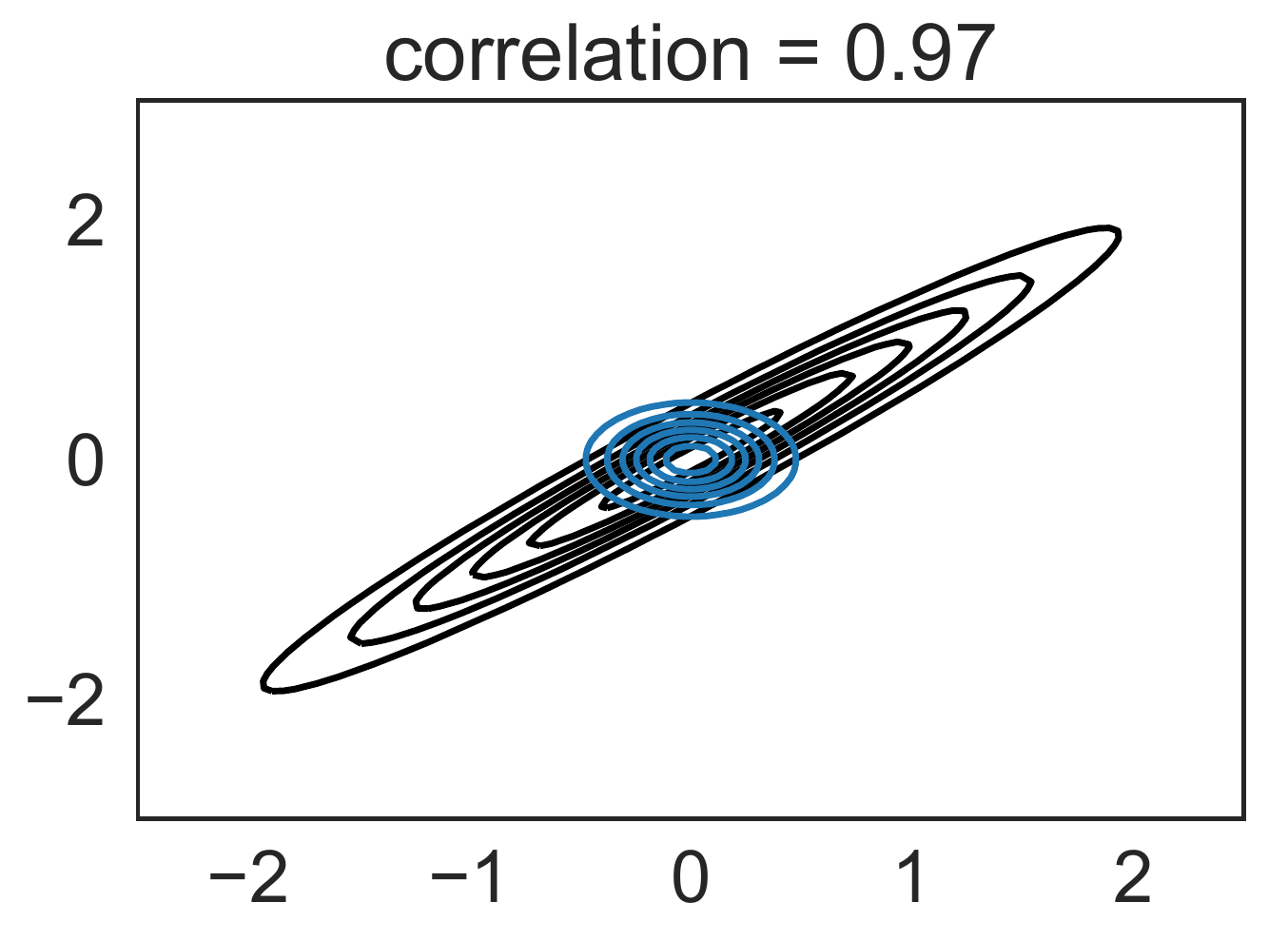}
\includegraphics[width=.32\textwidth]{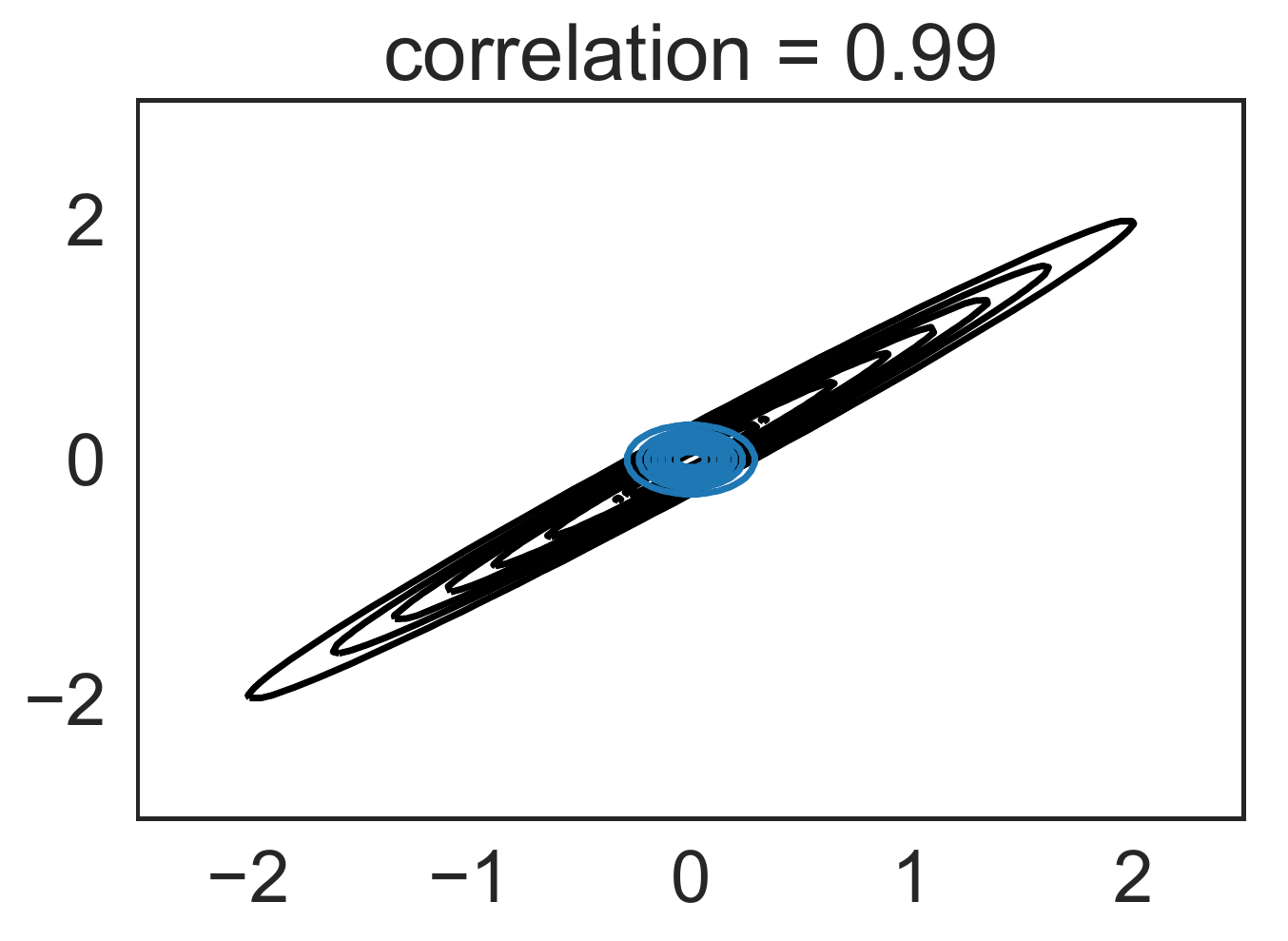}
\caption{}
\end{subfigure}
\caption{(a) KL divergence between optimal mean-field variational Gaussian approximation and Gaussian with the specified dimension (x-axis) and correlation (legend).
(b) 2-dimensional Gaussian with specified correlation (black) and mean-field variational Gaussian approximation (blue).}
\label{fig:kl-vb}
\end{center}
\end{figure*}

\section{KL divergence and variational inference} \label{sec:KL-discussion}

The KL divergence from a variational approximation to a posterior distribution can
vary greatly in size. A KL divergence value depends on the complexity of the posterior distribution, the flexibility of the variational family, and the 
optimization procedure employed. 
For example, \citet{Baque:2017} report KL divergences ranging from approximately 1 to nearly 500 depending on the model and 
variational family used. 

In \cref{fig:kl-vb}, we consider correlated Gaussians as the target distribution. We vary the dimension and pairwise correlation. And we plot the KL divergence from the optimal Gaussian mean-field 
variational approximation to the correlated Gaussian.
Gaussians offer a reasonable representation of many typical posteriors. They may arise, for instance,
(1) in conjugate linear regression or similar geometries or (2) due to the Bernstein--Von Mises theorem (i.e., Bayesian central limit theorem) 
\citep{vanderVaart:1998,Kleijn:2012}. 
Moreover, we might expect that observed KL divergence values given Gaussian targets may be smaller than we might observe for more complex targets.
Except in settings when the dimension and correlation are both fairly small, we observe KL divergence values greater than 1 in \cref{fig:kl-vb}.

\section{Transportation--divergence inequalities} \label{sec:intuition-and-details}

In this section we develop a deeper understanding of our bound in
\cref{cor:computable-bounds-simple} and its proof.
We show that our new theory -- including variations on 
our main bound \cref{cor:computable-bounds-simple} -- avoids the strong tail assumptions of
existing related work. 
Our results in this section are potentially of independent interest beyond Bayesian inference, so we
use the notation $\estmeas$ and $\mainmeas$ to represent two arbitrary
distributions; in the Bayesian setting, we would choose $\postdist =
\estmeas$ and $\approxdist = \mainmeas$.

There are a number of existing bounds on $\pwassSimple{p}{\estmeas}{\mainmeas}$
via $\kl{\estmeas}{\mainmeas}$, generally referred to as \emph{transportation--entropy inequalities}
(with reference to the other name for KL divergence, \emph{relative entropy}).
As discussed in \cref{sec:computablebounds}, these require a scale
parameter to modulate the bound. Existing bounds, however, are not sufficient
for our present purposes since they typically require impractically strong tail
assumptions. In particular, \cref{thm:EI2-WH1} in
\cref{sec:more-transport-entropy}, due to \citet{Bobkov:1999,Djellout:2004},
requires that $\mainmeas$ be 2-exponentially integrable and hence have lighter
or equal tails to a Gaussian. The following theorem -- which we use in 
\cref{cor:computable-bounds-simple} -- requires only exponential
tails to bound the 1-Wasserstein distance.
\begin{proposition}[{\citet[Corollary 2.3]{Bolley:2005}}] \label{prop:EIp-WH-type-bound}
Assume $\mainmeas$ is $p$-exponentially integrable for some $p \ge 1$.
Then for all $\estmeas \abscont \mainmeas$, 
\[
\pwassSimple{p}{\estmeas}{\mainmeas} \le \EIC{p}{\mainmeas} \left[\kl{\estmeas}{\mainmeas}^{\frac{1}{p}} + \{\kl{\estmeas}{\mainmeas}/2\}^{\frac{1}{2p}}\right]. 
\]
\end{proposition}
However, many posteriors of interest have much heavier tails -- often with at most
polynomial decay.  For example, neither inverse Gamma distributions nor 
$t$-distributions with $h < \infty$ degrees of freedom have exponential tails.
Moreover, bounding the 2-Wasserstein distance with \cref{prop:EIp-WH-type-bound} 
requires the problematic Gaussian tails assumption.

In contrast to these past results, our work provides bounds on Wasserstein
distances assuming only polynomial tail decay. We achieve these bounds by
incorporating more general $\alpha$-divergences; we call 
these new bounds \emph{transportation--divergence inequalities}. 
For example, \cref{thm:polynomial-moments-simple}
is a particularly simple bound on the $p$-Wasserstein distance in terms of just the
2-divergence when $\mainmeas$ has finite $(2p)$th moment.
We use this result, together with
\cref{lem:divergence-elbo-cubo-bounds} and \cref{prop:EIp-WH-type-bound}, to
prove \cref{cor:computable-bounds-simple} above.
\begin{proposition}\label{thm:polynomial-moments-simple}
Assume $\mainmeas$ is $2p$-polynomially integrable for some $p \ge 1$.
Then for all $\estmeas \abscont \mainmeas$, 
\[
\pwassSimple{p}{\estmeas}{\mainmeas} \leq \PIC{2p}{\mainmeas} \left[\exp\left\{\alphaDiv{2}{\estmeas}{\mainmeas}\right\} - 1\right]^{\frac{1}{2p}}.
\]
\end{proposition}

Next, we show how to achieve tighter bounds than \cref{thm:polynomial-moments-simple} via two additional novel transportation--entropy inequalities;
these can be combined with \cref{lem:divergence-elbo-cubo-bounds}
to arrive at results like \cref{cor:computable-bounds-simple},
at the price of additional complexity in the statements of the bounds.
 
Our first result offers a better dependence on moments in the exponential tails case by using both KL divergence and $\alpha$-divergence (cf.\ just KL divergence in \cref{prop:EIp-WH-type-bound}); however, the bound is more complex than \cref{prop:EIp-WH-type-bound}.
In particular, if $\mainmeas$ has exponential tails and we can bound the $\alpha$-divergence for any $\alpha > 1$,
then we can bound the 2-Wasserstein distance.

\begin{theorem}\label{thm:sqrt-EI}
Assume $\mainmeas \in \EI{p/2}{\eps}$ (\cref{def:expo_integ}) for some $p \ge 1$ and $\eps > 0$ and let $\EIconst{p}{\mainmeas}{\eps}$
be defined as in \cref{def:exp_int_constant}.
Let
\[
C(\alpha, \estmeas, \mainmeas) 
&\defined \inf_{\eps > 0} \Bigg\lbrace\frac{6}{\epsilon^2}\bigg[\left(\frac{3\alpha}{\alpha-1}\right)^2+6+2\EIconst{p/2}{\mainmeas}{\eps}^2 %
+\alphaDiv{\alpha}{\estmeas}{\mainmeas}^2\bigg]\Bigg\rbrace^{1/p}.
\]
Then for $\alpha > 1$ and $\estmeas \abscont \mainmeas$, 
\[
\pwassSimple{p}{\estmeas}{\mainmeas}
\leq C(\alpha, \estmeas, \mainmeas)\kl{\estmeas}{\mainmeas}^{\frac{1}{2p}}.
\]
\end{theorem}

Our second result requires only that $\mainmeas$ have a finite $(2pq)$th moment in order to bound 
the $p$-Wasserstein distance by the relative entropy and the $\alpha$-divergence.
Here, $q = q(\alpha) \defined \alpha/(\alpha - 1)$ is the conjugate exponent for $\alpha$. 
Thus, this result has a higher moment dependence than our \cref{thm:polynomial-moments-simple}, but it uses the $\alpha$-divergence with $\alpha < 2$ 
(cf.\ $\alpha = 2$ in \cref{thm:polynomial-moments-simple}) and thereby could produce tighter bounds.
\begin{theorem}\label{thm:polynomial-moments}
Fix $p \geq 1$ and $\alpha > 1$, and let $q = q(\alpha) \defined \alpha/(\alpha - 1)$.
Assume that $\mainmeas$ is $2pq$-polynomially integrable, as defined in \cref{sec:computablebounds}, and let 
\[
C(\alpha, \estmeas) \defined \inf_{\param'}&\left[\left(\int\metric(\param',\param)^{2p}\mainmeas(\dee\param)\right)^{1/2}
+\left(\frac{1}{2^{2q-2}q}\int \statictwonorm{\param' - \param}^{2pq}\mainmeas(\dee\param)
+ \frac{4e^{(\alpha-1)\alphaDiv{\alpha}{\estmeas}{\mainmeas}}}{\alpha}\right)^{1/2}\right]^{1/p}.
\]
Then for all $\estmeas \abscont \mainmeas$, 
\[
\pwassSimple{p}{\estmeas}{\mainmeas}
\leq 2C(\alpha, \estmeas)\;\kl{\estmeas}{\mainmeas}^{\frac{1}{2p}}.
\]
\end{theorem}

\section{Transportation--entropy inequality results} \label{sec:more-transport-entropy}

Classical transportation--entropy inequalities take the following form.
\begin{definition} \label{def:transport-entropy}
For $p \ge 1$ and $\rho > 0$, the distribution $\mainmeas$ satisfies a \emph{$p$-transportation--entropy (or $p$-Talagrand) 
inequality with constant $\rho$} (denoted $\mainmeas \in \WHsimple{p}{\rho}$) if for all $\estmeas \abscont \mainmeas$, 
\[
\pwassSimple{p}{\estmeas}{\mainmeas} \le \left\{\frac{2\,\kl{\estmeas}{\mainmeas}}{\rho}\right\}^{1/2}. \label{eq:transport-entropy}
\]
\end{definition}
When $p=1$ there are interpretable necessary and sufficient conditions for $\mainmeas \in \WHsimple{1}{\rho}$. 
The most important is the $p$-exponential integrability condition, which we denote by $\mainmeas \in \EI{p}{\eps}$:
\begin{definition}[cf. \cref{sec:computablebounds}]\label{def:expo_integ}
For $p \ge 1$ and $\epsilon>0$ the distribution $\mainmeas$ is $p$-exponentially integrable with parameter $\epsilon$ (denoted $\mainmeas \in \EI{p}{\eps}$) if
\[\inf_{\theta_0}\left[\int e^{\eps\staticnorm{\param - \param_0}_{2}^{p}}\mainmeas(\dee \param)\right]< \infty.\]
\end{definition} 
In particular, the following result shows that $\mainmeas$ satisfies a $1$-transportation--entropy inequality if and only if it has Gaussian tails. 
Moreover, the $\epsilon$ parameter in the corresponding 2-exponential integrability condition essentially determines
the precision of the transportation--entropy inequality. 

\begin{theorem}[{\citet[Theorem 3.1]{Bobkov:1999}, \citet[Theorem 2.3]{Djellout:2004}}] \label{thm:EI2-WH1}
The following conditions are equivalent:
\benum
\item For some $\rho > 0$, $\mainmeas \in \WHsimple{1}{\rho}$, as defined in \cref{def:transport-entropy}.
\item For some $\eps > 0$, $\mainmeas \in \EI{2}{\eps}$, as defined in \cref{def:expo_integ}.
\item \label{item:subgauss} There exists a constant $c > 0$ such that for every $\phi : \reals^{d} \to \reals$ with $\norm{\phi}_{L} \le 1$  (where $\norm{\cdot}_{L}$ denotes the Lipschitz constant)
and every $t \in \reals$,
\[
\mainmeas(e^{t\phi})
\le e^{ct^{2}}. 
\]
\eenum
Moreover, we may take $c = \rho^{-1}$ and
\[
c \le \frac{2}{\eps} \sup_{k \ge 1} \left\{\frac{(k!)^{2}}{(2k)!}\int \int e^{\eps\,\twonorm{\param - \param'}^{2}}\mainmeas(\dee \param)\mainmeas(\dee \param')\right\}^{1/k}.
\]
\end{theorem}
\begin{remark} \label{rmk:HW1-sub-Gaussian}
Let $\paramrv \dist \mainmeas$ and for a Lipschitz function $\phi : \reals^{d} \to \reals$, let $c_{\phi} \defined 2\norm{\phi}_{L}^{2}c$.
Condition (3) implies that the random variable $\phi(\paramrv)$ is $c_{\phi}$-sub-Gaussian~\citep[\S2.3]{Boucheron:2013}.
In particular, we have the concentration inequality 
\[
\Pr\{\phi(\paramrv) - \mainmeas(\phi) > t\} \le e^{-\frac{t^{2}}{2c_{\phi}}}.
\]
\end{remark}

The implication (2) $\implies$ (1) from \cref{thm:EI2-WH1} can be generalized to cover $p > 1$.
\begin{definition}\label{def:exp_int_constant}
For $p\ge 1$, the \emph{optimal $p$-exponential integrability constant} is given by
\[\EIconst{2p}{\mainmeas}{\eps}\defined \inf_{\param'}\log\int e^{\epsilon\twonorm{\param-\param'}^p}\mainmeas(\dee \param).
\]
\end{definition}
\begin{proposition}[{\citet[Corollary 2.4]{Bolley:2005}}] \label{prop:EIp-W2pH}
Assume $\mainmeas \in \EI{2p}{\eps}$ (\cref{def:expo_integ}) for some $p \ge 1$ and $\eps > 0$ and let 
\[
C \defined 2 \inf_{\eps > 0}\left[\frac{1}{2\eps}\left\{1 + \EIconst{2p}{\mainmeas}{\eps}\right\}\right]^{\frac{1}{2p}} < \infty,
\]
for $\EIconst{2p}{\mainmeas}{\eps}$ defined in \cref{def:exp_int_constant}.
Then for all $\estmeas \abscont \mainmeas$, 
\[
\pwassSimple{p}{\estmeas}{\mainmeas} \le C\,\kl{\estmeas}{\mainmeas}^{\frac{1}{2p}}.
\]
\end{proposition}

If one can establish that $\mainmeas \in \WHsimple{p}{\rho}$, then the pushforward measure under a Lipschitz transformation
also satisfies a $p$-transportation--entropy inequality. 
\begin{lemma}
Assume that for some $\rho > 0$, $\mainmeas \in \WHsimple{p}{\rho}$, and that $\Psi : \reals^{d} \to \reals^{d}$ is 
$L$-Lipschitz; i.e.,
\[
\twonorm{\Psi(\param) -  \Psi(\param')} \le L\,\twonorm{\param - \param'} \quad \param,\param' \in \reals^{d}.
\]
Then $\Psi\#\mainmeas \in \WHsimple{p}{\rho/L^{2}}$.
\end{lemma}

We close with the interesting connection that $\mainmeas \in \WHsimple{2}{\rho}$ is equivalent to $\mainmeas$ satisfying a dimension-free
Gaussian concentration inequality (cf.\ \cref{rmk:HW1-sub-Gaussian}). 
While the concentration condition is not necessarily easy to check, it does offer insight into what it means for $\mainmeas \in \WHsimple{2}{\rho}$.
\begin{theorem}[{\citet[Theorem 1.3]{Gozlan:2009}}]
For a set $A \subseteq (\reals^{d})^{n}$, let $A^{t} \defined \{ \param \in (\reals^{d})^{n} \given \exists \param' \in A : \sum_{i=1}^{n}\twonorm{\param_{i}-  \param'_{i}}^{2} \le t^{2} \}$. 
The following conditions are equivalent:
\benum
\item For some $\rho > 0$, $\mainmeas \in \WHsimple{2}{\rho}$.
\item There exist $a > 0, b > 0$ such that for all $n \in \nats$ and measurable $A \subseteq E^{n}$,
with $\mainmeas^{\otimes n}(A) \ge 1/2$, the probability measure $\mainmeas^{\otimes n}$ satisfies
\[
\mainmeas^{\otimes n}(A^{t}) \ge 1 - b e^{-at^{2}}.
\]
\eenum
\end{theorem}

\input{proofs}

\end{adjustwidth}

%% file: proofs.tex
\section{Proofs} \label{sec:proofs}

\subsection{Proofs of \cref{ex:KL-divergence-problems1,ex:KL-divergence-problems2,ex:2-div-problems}} \label{sec:KL-divergence-problems-details}

For \cref{ex:KL-divergence-problems1}(A), we let $\approxdist = \distWeibull(k/2, 1)$ and $\postdist = \distWeibull(k, 1)$.
Let $\gamma$ be the Euler-Mascheroni constant and $\Gamma$ be the gamma function.
We obtain~\citep{Bauckhage:2013}
\[
\kl{\approxdist}{\postdist} = -\log(2)+\gamma+\Gamma\left(3\right)-1 < 0.9.
\]
Using the well-known formulas for the mean and variance of the Weibull distribution, we have
$\mean{\approxdist}=\Gamma(1+2/k)$, $\mean{\postdist}=\Gamma(1+1/k)$, 
and $\var{\postdist}=\Gamma(1+2/k) - \{\Gamma(1+1/k)\}^2$. 
Hence,  $\lim_{k\searrow 0}(\mean{\approxdist}-\mean{\postdist})^2/\var{\postdist} =\infty$. 

For \cref{ex:KL-divergence-problems1}(B), let $\approxdist = \distWeibull(k, 1)$ and $\postdist = \distWeibull(k/2, 1)$.
We obtain
\[
\kl{{\approxdist}}{{\postdist}}=\log(2)- \gamma/2 +\Gamma(3/2) -1<0.3.
\]
By the same argument as above,
$\lim_{k\searrow 0}(\mean{{\approxdist}}-\mean{{\postdist}})^2/\var{{\approxdist}}=\infty$.

For \cref{ex:KL-divergence-problems2}, we let $\approxdist$ be standard normal and $\postdist=\mcT_{h}$ be a standard $t$-distribution with $h$ degrees of freedom. Let $\paramrv\sim\approxdist$. It is straightforward to show that
\[
\kl{\approxdist}{\postdist} 
&= \log[\Gamma(h/2)h^{1/2}/\Gamma\{(h + 1)/2\}]
 - 0.5 \log(2e)  + 0.5(h + 1)\EE\left\{\log\left(1 + \paramrv^{2}/h\right)\right\}. 
\]
For $h = 2$, this quantity can be numerically evaluated and is less than $0.12$.
By continuity of the function $h\mapsto\kl{\approxdist}{\mcT_{h}}$, there exists some $\epsilon>0$, such that for all $h\in[2,2+\epsilon)$, $\kl{\approxdist}{\mcT_{h}}<0.12$. Finally, we observe that $\lim_{h \searrow 2}\var{\mcT_{h}} = \infty$.

For \cref{ex:2-div-problems}, we choose $\postdist = \distWeibull(k, 1)$ and $\approxdist = \distWeibull(k/2, 1)$ for $k>0$. Note that $\lim_{k\downarrow 0}\frac{\sigma_{\approxdist}^2}{\sigma_{\postdist}^2}=\infty$ and $\lim_{k\downarrow 0}(\mean{{\approxdist}}-\mean{{\postdist}})^2/\var{{\postdist}}=\infty.$ On the other hand, letting $f_{\approxdist}$ and $f_{\postdist}$ be the densities of $\approxdist$ and $\postdist$, respectively, we have
\[
\lefteqn{\int_0^{\infty} \left(f_{\postdist}(x)\right)^{2}\left(f_{\approxdist}(x)\right)^{-1}\dee x}\\
&=2 k \int_0^{\infty} x^{3k/2-1}\exp\left(-2 x^{k}+x^{k/2}\right)\dee x\\
&\stackrel{y=x^{k/2}}=4\int_0^{\infty}y^{2}\exp\left(-2 y^2+ y\right) \dee y<1.47752
\]
and so
\[\alphaDiv{2}{\postdist}{\approxdist}=\log\int_0^{\infty} \left(f_{\postdist}(x)\right)^{2}\left(f_{\approxdist}(x)\right)^{-1}\dee x<0.391.\]
Therefore, for any $t>0$, there exist two distributions $\approxdist$ and $\postdist$ with $\alphaDiv{2}{\postdist}{\approxdist}$ bounded by $0.391$ yet such that $\sigma_{\approxdist}^2\ge t\sigma_{\postdist}^2$ and $(\mean{{\approxdist}}-\mean{{\postdist}})^2\ge t\var{\postdist}$.

\subsection{Proof of \cref{ex:no-problems1}}
Let $\alpha>1$. Then $\estmeas$ and $\mainmeas$ have the following densities w.r.t. the Lebesgue measure
\[ 
&f_{\estmeas}(x)=
\frac{k}{2}x^{k/2-1}e^{-x^{k/2}}\mathbb{I}[x\geq 0],\,
f_{\mainmeas}(x)=
kx^{k-1}e^{-x^{k}}\mathbb{I}[x\geq 0]\]
and
\[\alphaDiv{\alpha}{\estmeas}{\mainmeas}=\frac{1}{\alpha-1}\log\int_0^{\infty} \left(f_{\estmeas}(x)\right)^{\alpha}\left(f_{\mainmeas}(x)\right)^{1-\alpha}\dee x.\]
Note that
\[
&\lefteqn{\int_0^{\infty} \left(f_{\estmeas}(x)\right)^{\alpha}\left(f_{\mainmeas}(x)\right)^{1-\alpha}\dee x}\\
&=\frac{k}{2^{\alpha}}\int_0^{\infty} x^{k-1-k\alpha/2}\exp\left(-\alpha x^{k/2}+(\alpha-1)x^k\right)\dee x\\
&\stackrel{y=x^{k/2}}=\frac{1}{2^{\alpha-1}}\int_0^{\infty} y^{1-\alpha}\exp\left(-\alpha y + (\alpha-1)y^2\right) \dee y\\
&=\infty.
\]
Therefore, for $\alpha>1$, $\alphaDiv{\alpha}{\estmeas}{\mainmeas}=\infty$.

Similarly, 
\[
\lefteqn{\int_0^{\infty} \left(f_{\mainmeas}(x)\right)^{\alpha}\left(f_{\estmeas}(x)\right)^{1-\alpha}\dee x}\\
&=2^{\alpha-1} k \int_0^{\infty} x^{k\alpha/2+k/2-1}\exp\left(-\alpha x^{k}-x^{k/2}+\alpha x^{k/2}\right)\dee x\\
&\stackrel{y=x^{k/2}}=2^{\alpha}\int_0^{\infty}y^{\alpha}\exp\left(-\alpha y^2+ (\alpha-1)y\right) \dee y<\infty.
\]
Therefore, for $\alpha>1$, $\alphaDiv{\alpha}{\mainmeas}{\estmeas}<\infty$. However, 
$$\int_0^{\infty}|x-x'|^2\estmeas(\dee x)=\Gamma\left(1+\frac{4}{k}\right)-2x'\Gamma\left(1+\frac{2}{k}\right)+(x')^2.$$
Minimizing this over $x'$ gives us that the minimum is achieved at $x'=\Gamma\left(1+\frac{2}{k}\right)$. But
\[
&\lim_{k\searrow 0}\left[\Gamma\left(1+\frac{4}{k}\right)-2\left(\Gamma\left(1+\frac{2}{k}\right)\right)^2+\left(\Gamma\left(1+\frac{2}{k}\right)\right)^2\right]
=\infty
\]
and so $\PIC{p}{\estmeas} \nearrow \infty$ as $k\searrow 0$.

\subsection{Proof of \cref{ex:no-problems2}}
Letting $f_{\estmeas}$ and $f_{\mainmeas}$ be the corresponding densities, we have
\[
\lefteqn{\int_{-\infty}^{\infty}\left(f_{\estmeas}\right)^{\alpha}\left(f_{\mainmeas}\right)^{1-\alpha}\dee x}\\
&=\frac{1}{(2\pi)^{\alpha/2}}\left(\frac{\Gamma\left((h+1)/2\right)}{\sqrt{h\pi}\Gamma(h/2)}\right)^{1-\alpha}\int_{-\infty}^{\infty}e^{-\alpha x^2/2}\left(1+\frac{x^2}{h}\right)^{(h+1)(\alpha-1)/2}\dee x.\\
&\leq \frac{1}{(2\pi)^{\alpha/2}}\left(\frac{\Gamma\left((h+1)/2\right)}{\sqrt{h\pi}\Gamma(h/2)}\right)^{1-\alpha}\int_{-\infty}^{\infty}e^{-x^2/2}\dee x<\infty,
\]
because, for $h\geq 2$, $\left(1+\frac{x^2}{h}\right)^{(h+1)(\alpha-1)/2}\leq e^{(\alpha-1)x^2/2}$. Therefore, $\alphaDiv{\alpha}{\estmeas}{\mainmeas}<\infty$. However, for $h>2$,
\[
\int_{-\infty}^{\infty}|x-x'|^2\mainmeas(\dee x)=\frac{h}{h-2}+(x')^2\geq \frac{h}{h-2}\xrightarrow{h\searrow 2}\infty
\]
and $\PIC{p}{\mainmeas} \nearrow \infty$ as $h\searrow 2$.

\subsection{Proof of \cref{lem:scale-invariance}}\label{sec:proof_scale_inv}
First assume that $\estmeas$ and $\mainmeas$ have densities $f_{\estmeas}$ and $f_{\mainmeas}$ with respect to Lebesgue measure. Note that the densities of the pushforward measures $T\#\estmeas$ and $T\#\mainmeas$ are given by 
\[x\mapsto f_{\estmeas}\circ T^{-1}(x)\left|\text{det}J_xT^{-1}(x)\right|\quad\text{and}\quad x\mapsto f_{\mainmeas}\circ T^{-1}(x)\left|\text{det}J_xT^{-1}(x)\right|,\] respectively, where $J_x$ denotes the Jacobian. Therefore, for any $\alpha>0$,
\[ \int \left(\der{\left(T\#\estmeas\right)}{\left(T\#\mainmeas\right)}\right)^{\alpha}\dee\left(T\#\mainmeas\right)
&=\int \left(\frac{f_{\estmeas}\circ T^{-1}(x)\left|\text{det}J_xT^{-1}(x)\right|}{f_{\mainmeas}\circ T^{-1}(x)\left|\text{det}J_xT^{-1}(x)\right|}\right)^{\alpha}f_{\mainmeas}\circ T^{-1}(x)\left|\text{det}J_xT^{-1}(x)\right|\dee x\\
&=\int \left(\frac{f_{\estmeas}\circ T^{-1}(x)}{f_{\mainmeas}\circ T^{-1}(x)}\right)^{\alpha}f_{\mainmeas}\circ T^{-1}(x)\left|\text{det}J_xT^{-1}(x)\right|\dee x\\
&\stackrel{y=T^{-1}(x)}=\int \left(\frac{f_{\estmeas}(y)}{f_{\mainmeas}(y)}\right)^{\alpha}f_{\mainmeas}(y)\dee y\\
&=\int \left(\der{\estmeas}{\mainmeas}\right)^{\alpha}\dee \mainmeas.
\]
and so, for $\alpha\neq 1$, $\alphaDiv{\alpha}{\estmeas}{\mainmeas} = \alphaDiv{\alpha}{T \# \estmeas}{T \# \mainmeas}$. Similarly,
\[ \int \log\left(\der{\left(T\#\estmeas\right)}{\left(T\#\mainmeas\right)}\right)\dee\left(T\#\estmeas\right)
&=\int \log\left(\frac{f_{\estmeas}\circ T^{-1}(x)}{f_{\mainmeas}\circ T^{-1}(x)}\right)f_{\estmeas}\circ T^{-1}(x)\left|\text{det}J_xT^{-1}(x)\right|\dee x\\
&=\int \log\left(\frac{f_{\estmeas}(y)}{f_{\mainmeas}(y)}\right)f_{\estmeas}(y)\dee y\\
&=\int \log\left(\der{\estmeas}{\mainmeas}\right)\dee \mainmeas.\]
and so $\alphaDiv{1}{T \# \estmeas}{T \# \mainmeas}=\kl{T\#\vardist}{T\#\postdist}=\kl{\vardist}{\postdist} =\alphaDiv{1}{\estmeas}{\mainmeas}$.

More generally, without assuming that $\estmeas$ and $\mainmeas$ are absolutely continuous with respect to Lebesgue measure, we note that if $\estmeas\ll \mainmeas$ then 
\[\der{\left(T\#\estmeas\right)}{\left(T\#\mainmeas\right)}=\der{\estmeas}{\mainmeas}\circ T^{-1}.\label{eq:radon}\] Indeed, for any measurable set $A$, we have
\[\int_A\der{\estmeas}{\mainmeas}\circ T^{-1}\,\dee\left(T\#\mainmeas\right)=\int_{T^{-1}(A)}\der{\estmeas}{\mainmeas}d\mainmeas=\left(T\#\estmeas\right) (A).
\]
Using \cref{eq:radon} and the fact that $T$ is bijective, we have that
\[ \int \left(\der{\left(T\#\estmeas\right)}{\left(T\#\mainmeas\right)}\right)^{\alpha}\dee\left(T\#\mainmeas\right)=\int \left(\der{\estmeas}{\mainmeas}\circ T^{-1}\right)^{\alpha}\dee\left(T\#\mainmeas\right)=\int \left(\der{\estmeas}{\mainmeas}\right)^{\alpha}\dee\mainmeas.\label{eq:radon1}\]
Similarly,
\[ \int \left(\der{\log\left(T\#\estmeas\right)}{\left(T\#\mainmeas\right)}\right)\dee\left(T\#\estmeas\right)=\int \log\left(\der{\estmeas}{\mainmeas}\circ T^{-1}\right)\dee\left(T\#\estmeas\right)=\int \log\left(\der{\estmeas}{\mainmeas}\right)\dee\estmeas.\label{eq:radon2}\]
\cref{eq:radon1} and \cref{eq:radon2} prove that $\alphaDiv{\alpha}{\estmeas}{\mainmeas} = \alphaDiv{\alpha}{T \# \estmeas}{T \# \mainmeas}$ for any $\alpha>0.$

\subsection{Proof of \cref{thm:Wasserstein-moment-bounds}}

We begin by considering the case $d=1$, dropping the component indexes from our notation. 
\begin{theorem} \label{thm:one-dim-Wasserstein-moment-bounds}
Assume $d = 1$. 
If $\pwassSimple{1}{\mainmeas}{\estmeas} \le \veps$, then 
$
|\mean{\mainmeas} - \mean{\estmeas}| \le \veps 
$
and
$
|\MAD{\mainmeas} - \MAD{\estmeas}| \le 2\veps. %
$
On the other hand, if $\pwassSimple{2}{\mainmeas}{\estmeas} \le \veps$, then 
$\pwassSimple{1}{\mainmeas}{\estmeas} \le \veps$,
\[
|\std{\mainmeas} - \std{\estmeas}| \le \veps, \label{eq:stdev-error-bound}
\]
and
\[
|\var{\mainmeas} - \var{\estmeas}| 
&\le 2\min(\std{\mainmeas}, \std{\estmeas})\veps + 2\veps^{2} .\label{eq:var-error-bound}
\]
\end{theorem}

The proof of \cref{thm:one-dim-Wasserstein-moment-bounds} is deferred to the next section. 
To generalize to the case of $d > 1$, for a random variable $\paramrv \dist \eta$ on $\reals^{d}$ with
distribution $\eta$ and any vector $v \in \reals^{d}$, let $\mean{\eta,v} = \EE(v^{\top}\paramrv)$,
$\var{\eta,v} = \EE\{(v^{\top}\paramrv -\mean{\eta,v})^{2}\}$,  and
$\MAD{\eta,v} = \EE(|v^{\top}\paramrv - \mean{\eta,v}|)$.

\begin{corollary} \label{cor:multidim-Wasserstein-moment-bounds}
Let $v \in \reals^{d}$ satisfy $\twonorm{v} \le 1$. 
If $\pwassSimple{1}{\mainmeas}{\estmeas} \le \veps$ then 
$
|\mean{\mainmeas,v} - \mean{\estmeas,v}| \le \veps
$
and
$
|\MAD{\mainmeas,v} - \MAD{\estmeas,v}| \le 2\veps. \label{eq:v-mean-and-v-MAD-bounds}
$
On the other hand, if $\pwassSimple{2}{\mainmeas}{\estmeas} \le \veps$, then
\[
|\std{\mainmeas,v} - \std{\estmeas,v}| 
\le \veps, \qquad
|\var{\mainmeas,v} - \var{\estmeas,v}| 
	\le 2\min(\std{\mainmeas,v}, \std{\estmeas,v})\veps + 2\veps^{2}.
\]
\end{corollary}
\bprf
Let $\paramrv \dist \mainmeas$, let $\paramrv_{v} = v^{\top}\paramrv$ and let $\mainmeas_{v}$ denote the 
distribution of $\paramrv_{v}$. 
Define $\hat{\paramrv}$, $\hat{\paramrv}_{v}$, and $\estmeas_{v}$ analogously in terms of $\estmeas$. 
By the Cauchy-Schwarz inequality and the assumption that $\twonorm{v} \le 1$, we have that, for any $p\geq 1$,
\[
\EE(|\paramrv_{v} - \hat{\paramrv}_{v}|^{p})
= \EE(|v^{\top}\paramrv - v^{\top}\hat{\paramrv}|^{p})
\le \EE(\statictwonorm{\paramrv - \hat{\paramrv}}^{p}).
\]
Hence $\pwassSimple{p}{\mainmeas_{v}}{\estmeas_{v}} \le \pwassSimple{p}{\mainmeas}{\estmeas}$. 
The corollary now follows from \cref{thm:one-dim-Wasserstein-moment-bounds}. 
\eprf
\begin{lemma} \label{lem:sup-v-equalities}
For probability measures $\xi, \mainmeas, \estmeas$, we have  
$\twonorm{\mean{\mainmeas} - \mean{\estmeas}} = \sup_{\twonorm{v} \le 1}|\mean{\mainmeas,v} - \mean{\estmeas,v}|$, 
$\twonorm{\Sigma_{\xi}} = \sup_{\twonorm{v} \le 1} \var{\xi,v}$, and  
$\twonorm{\Sigma_{\mainmeas} - \Sigma_{\estmeas}} = \sup_{\twonorm{v} \le 1} |\var{\mainmeas,v} - \var{\estmeas,v}|$. 
\end{lemma}
\bprf
The first result follows since $\mean{\mainmeas,v} - \mean{\estmeas,v} = v^{\top}(\mean{\mainmeas} - \mean{\estmeas})$
and for any $w \in \reals^{d}$, $\sup_{\twonorm{v} \le 1} v^{\top}w = \twonorm{w}$. 
For the second result, since $\Sigma_{\xi}$ is positive semi-definite,
\[
\twonorm{\Sigma_{\xi}}
&= \sup_{\twonorm{v} \le 1} v^{\top}\Sigma_{\xi}v  
= \sup_{\twonorm{v} \le 1}\EE\{v^{\top}(\Xvec - \meanvec{\xi})(\Xvec - \meanvec{\xi})^{\top}v\} 
= \sup_{\twonorm{v} \le 1}\var{\xi,v};
\]
The third result follows by an analogous argument.
\eprf

By taking $v = e_{i}$, the $i$th canonical basis vector of $\reals^{d}$, 
\cref{cor:multidim-Wasserstein-moment-bounds} implies the bounds in \cref{thm:Wasserstein-moment-bounds}
on $|\MAD{\mainmeas,i} - \MAD{\estmeas,i}|$ %
and $|\std{\mainmeas,i} - \std{\estmeas,i}|$.
\cref{cor:multidim-Wasserstein-moment-bounds,lem:sup-v-equalities} yield the bounds in \cref{thm:Wasserstein-moment-bounds}
on $\twonorm{\mean{\mainmeas} - \mean{\estmeas}}$ and $\twonorm{\Sigma_{\mainmeas} - \Sigma_{\estmeas}}$.

\subsection{Proof of \cref{thm:one-dim-Wasserstein-moment-bounds}}

Throughout we will always assume that $\paramrv \dist \mainmeas$ and $\hat{\paramrv} \dist \estmeas$
are distributed according to the optimal coupling for the $p$-Wasserstein distance under consideration.
We will also assume without loss of generality that $\mean{\mainmeas} = 0$ since if not
we could consider the random variables $\paramrv' = \paramrv - \mean{\mainmeas}$ and $\hat{\paramrv}' = \hat{\paramrv} - \mean{\mainmeas}$
instead. 

The 1-Wasserstein distance can be written as~\citep[Rmk.~6.5]{Villani:2009}
\[
\pwassSimple{1}{\mainmeas}{\estmeas}  = \sup_{\phi \st \norm{\phi}_{L} \le 1} |\mainmeas(\phi) - \estmeas(\phi)|.  \label{eq:1-Wasserstein-dual}
\]
By Jensen's inequality, 
\[
\pwassSimple{q}{\mainmeas}{\estmeas} 
	&\le \pwassSimple{p}{\mainmeas}{\estmeas} 
	\qquad (1 \le q \le p < \infty). \label{eq:p-Wasserstein-ordering}
\] 
\cref{eq:p-Wasserstein-ordering,eq:1-Wasserstein-dual} together imply that for any $p \ge 1$, 
if $\pwassSimple{p}{\mainmeas}{\estmeas} \le \veps$, then for any $L$-Lipschitz function $\phi$, 
$|\mainmeas(\phi) - \estmeas(\phi)| \le L\veps$.

Assume $\pwassSimple{1}{\mainmeas}{\estmeas} \le \veps$.
By \cref{eq:1-Wasserstein-dual}, for any Lipschitz function $\phi$,
\[
|\EE(\phi(\paramrv) - \phi(\hat{\paramrv}))| \le \veps\norm{\phi}_{L}. 
\]                                          
Hence, taking $\phi(t) = t$, we have that 
$
|\mean{\mainmeas} - \mean{\estmeas}| = |\mean{\estmeas}| \le \veps. 
$
For the mean absolute deviation, using the fact that $\phi(t) = |t|$
is 1-Lipschitz, we have 
\[
|\MAD{\mainmeas} - \MAD{\estmeas}|
&= |\EE(|\paramrv| - |\hat{\paramrv} - \mean{\estmeas}|)| 
\le |\EE(|\paramrv| - |\hat{\paramrv}|)| + |\mean{\estmeas}| 
\le 2\veps.
\]

Assume $\pwassSimple{2}{\mainmeas}{\estmeas} \le \veps$.
By Jensen's inequality 
$\pwassSimple{1}{\mainmeas}{\estmeas} \le \veps$ as well. 
Let $\varsigma_{\mainmeas}^{2} = \EE(\paramrv^{2}) = \var{\mainmeas}$ and $\varsigma_{\estmeas}^{2} = \EE(\hat{\paramrv}^{2})$.
It follows from the Cauchy-Schwarz inequality that 
$\left(\varsigma_{\mainmeas}-\varsigma_{\estmeas}\right)^2\leq \EE\left((\paramrv-\hat{\paramrv})^2\right)=\left(\pwassSimple{2}{\mainmeas}{\estmeas}\right)^2$
and so
\[
|\varsigma_{\mainmeas} - \varsigma_{\estmeas}| &\le \veps.  \label{eq:bar-sigma-diff}
\]
Using \cref{eq:bar-sigma-diff}, we also have
\[
|\var{\mainmeas} - \var{\estmeas}|
    &= |\varsigma_{\mainmeas}^{2} - \varsigma_{\estmeas}^{2} + \mean{\estmeas}^{2}| 
    \le |\varsigma_{\mainmeas}^{2} - \varsigma_{\estmeas}^{2}| + |\mean{\estmeas}^{2}|
    \le \veps (\varsigma_{\mainmeas} + \varsigma_{\estmeas}) + \veps^{2} \label{eq:var-error-preliminary} 
\]
 Moreover, note that
 \[
\left( \pwassSimple{2}{\mainmeas}{\estmeas}\right)^2=\EE\left(\paramrv^2\right)+\EE\left(\hat{\paramrv}^2\right)-2\EE\left(\paramrv\hat{\paramrv}\right)=\var{\mainmeas} + \var{\estmeas}+\mean{\estmeas}^2-2\std{\paramrv\hat{\paramrv}}.
 \]
 From Cauchy-Schwarz, $\std{\paramrv\hat{\paramrv}}\leq \std{\mainmeas}\std{\estmeas}$, so that
$ \left(\std{\mainmeas}-\std{\estmeas}\right)^2+\mean{\estmeas}^2\leq  \left(\pwassSimple{2}{\mainmeas}{\estmeas}\right)^2$
 and so
\[
|\std{\mainmeas} - \std{\estmeas}| 
    &\le \veps. \label{eq:stdev-error-preliminary}
\]
Starting from \cref{eq:var-error-preliminary} and using \cref{eq:bar-sigma-diff}, we have
\[
|\var{\mainmeas} - \var{\estmeas}| 
&\le \veps (\varsigma_{\mainmeas} + \varsigma_{\estmeas}) + \veps^{2}  
\le \veps(2\varsigma_{\mainmeas}+\veps)+\veps^2  = \veps (2\std{\mainmeas} + \veps) + \veps^{2} 
= 2\,\std{\mainmeas}\veps + 2\veps^{2}.
\]

\subsection{Proof of \cref{prop:predictive-Wasserstein}}

Let $\gamma_{\theta,\theta'}^{*}$ denote the optimal $p$-Wasserstein coupling for $f(\cdot \given \param)$ and $f(\cdot \given \param')$.
Then we have
\[
\pwassSimple{p}{\hat{\mu}}{\mu}^{p} 
&=  \inf_{\coupling \in \couplings{\hat{\mu}}{\mu}} \left\{ \int \staticnorm{z - z'}_{2}^{p} \coupling(\dee z, \dee z') \right\} \\
&\le  \inf_{\coupling \in \couplings{\approxdist}{\postdist}} \left\{ \int\int \staticnorm{z - z'}_{2}^{p}\gamma_{\theta,\theta'}^{*}(\dee z, \dee z') \coupling(\dee \param, \dee \param')  \right\} \\
&=  \inf_{\coupling \in \couplings{\approxdist}{\postdist}} \left\{ \int\pwassSimple{p}{f(\cdot \given \param)}{f(\cdot \given \param')}^{p}  \coupling(\dee \param, \dee \param')  \right\} \\
&\le  \inf_{\coupling \in \couplings{\approxdist}{\postdist}} \left\{ c_{f}^{p}\int \twonorm{\param - \param'}^{p}  \coupling(\dee \param, \dee \param')  \right\} \\
&= c_{f}^{p} \pwassSimple{p}{\approxdist}{\postdist}^{p}. 
\]

\subsection{Proof of \cref{lem:divergence-elbo-cubo-bounds}} \label{sec:elbo_cubo_bounds_proof}
\bprf
First, note that the $\elbo{\vardist}$ provides a lower 
bound for $\log \marginallik$ since $\kl{\vardist}{\postdist} \geq 0$:
\[
\elbo{\vardist} &\defined \int \log\left(\der{\postdist'}{\vardist} \right)\dee\vardist \\
&=\log\marginallik - \kl{\vardist}{\postdist}\le \log \marginallik. \label{eq:elbo-ml-bound}
\]

Second, Jensen's inequality implies that \cubo{\alpha}{\vardist} is an upper bound for $\log \marginallik$:
\[
\cubo{\alpha}{\vardist}
&\defined  \log \left\{\int \left(\der{\postdist'}{\vardist}\right)^{\alpha} \dee\vardist\right\}^{1/\alpha}  \\
&\ge  \log \left\{\int \der{\postdist'}{\vardist} \dee\vardist\right\} = \log \marginallik. \label{eq:cubo-ml-bound}
\]

The $\alpha$-divergence is monotone in $\alpha$, i.e., $\alpha \leq \alpha'$
implies that ${\alphaDiv{\alpha}{\postdist}{\approxdist}} \le
{\alphaDiv{\alpha'}{\postdist}{\approxdist}}$~ \citep{Cichocki:2010}.
Thus, by the definition of \cubo{\alpha}{\approxdist} and \cref{eq:elbo-ml-bound}, we have
\[
\kl{\postdist}{\approxdist} 
&=  \alphaDiv{1}{\postdist}{\approxdist} 
\le  \alphaDiv{\alpha}{\postdist}{\approxdist} \\
&= \frac{\alpha}{\alpha - 1} \left(\cubo{\alpha}{\approxdist} - \log\marginallik\right) \\
&\le \frac{\alpha}{\alpha - 1} \left(\cubo{\alpha}{\approxdist} - \elbo{\eta}\right).
\]
\eprf

\subsection{Proof of \cref{thm:sqrt-EI}} \label{sec:proof-of-sqrt-EI}

\begin{theorem}\label{thm:TV-sqrt-EI}
Let  $\varphi$ be a nonnegative measurable function on E and let $\delta>0$. Then we have
$$\|\varphi(\estmeas-\mainmeas)\|_{TV}\leq \left(27\left(\frac{1+\delta}{\delta}\right)^2+18+5\left(\log \int_Ee^{\sqrt{2\varphi}}\,\dee\mainmeas\right)^2+3\alphaDiv{1+\delta}{\estmeas}{\mainmeas}^2\right)\kl{\estmeas}{\mainmeas}^{1/2}.$$
\end{theorem}

Corollary \ref{thm:sqrt-EI} follows from Theorem \ref{thm:TV-sqrt-EI} and the fact that 
$$\pwassSimple{p}{\estmeas}{\mainmeas}^p \leq 2^{p-1}\|m(\paramrv',\cdot)^p(\estmeas-\mainmeas)\|_{TV},$$
proved, for instance, in \citet[Proposition 7.10]{Villani:2003}. Indeed, it suffices to use $\varphi=\frac{\epsilon^2}{2} m(\paramrv',\cdot)^p$ in Theorem \ref{thm:TV-sqrt-EI} to obtain the assertion.

\subsection{Proof of \cref{thm:TV-sqrt-EI}}

We first assume, without loss of generality, that $\estmeas$ is absolutely continuous with respect to $\mainmeas$, with density $f$. We set $u:=f-1$ so that
$$\estmeas=(1+u)\mainmeas$$
and note that $u\geq -1$ and $\int_Eu\dee\mainmeas=0$. We also define
$$h(v):=(1+v)\log(1+v)-v,\quad v\in[-1,+\infty)$$
so that
\begin{equation}\label{2}
\kl{\estmeas}{\mainmeas}=\int_Eh(u)\dee\mainmeas.
\end{equation}
We note that $h\geq 0$. We split the total variation in the following way:
\begin{equation}\label{1}
\int\varphi \,d|\estmeas-\mainmeas|=\int\varphi|u|\,\dee\mainmeas=\int_{\lbrace -1\leq u\leq 4\rbrace} \varphi|u|\,\dee\mainmeas+\int_{\lbrace u>4\rbrace} \varphi u\,\dee\mainmeas.
\end{equation}

\paragraph{First part of the proof.}

In the first part, the first term ($u\leq 4$) in (\ref{1}) is bounded. This part is an adaptation of the first part of the proof of \citet[Theorem 1]{Bolley:2005}.

By Cauchy-Schwarz,
$$\int_{\lbrace u\leq 4\rbrace} \varphi |u|\dee\mainmeas\leq \left(\int_{\lbrace u\leq 4\rbrace}\varphi^2\,\dee\mainmeas\right)^{1/2}\left(\int_{\lbrace u\leq 4\rbrace}u^2\,\dee\mainmeas\right)^{1/2}.$$
On the other hand, from the elementary inequality,
$$-1\leq v\leq 4\quad  \Longrightarrow\quad v^2\leq 4h(v)$$
(a consequence of the fact that $h(v)/v$ is nondecreasing), we deduce
$$\int_{\lbrace u\leq 4\rbrace} u^2\,\dee\mainmeas\leq 4\int_{\lbrace u\leq 4\rbrace}h(u)\,\dee\mainmeas.$$
Combining this with the nonnegativity of $h$ and (\ref{2}), we find that
\begin{equation}\label{3}
\int_{\lbrace u\leq 4\rbrace}\varphi|u|\,\dee\mainmeas\leq 2\left(\int_E\varphi^2\,\dee\mainmeas\right)^{1/2}\left(\int_Eh(u)\,\dee\mainmeas\right)^{1/2}=2\left(\int_E\varphi^2\,\dee\mainmeas\right)^{1/2}\kl{\estmeas}{\mainmeas}^{1/2}.
\end{equation}

Now, since the function $t\mapsto \exp(\sqrt{2}t^{1/4})$ in increasing and convex on $\left[\frac{9^2}{2^2},+\infty\right)$, we can write
\[\begin{aligned}
&\exp\left[\sqrt{2}\left(\int_E\varphi^2\, \dee\mainmeas\right)^{1/4}\right]\\
\leq&\exp\left[\sqrt{2}\left(\int_E\left(\varphi+\frac{9}{2}\right)^2\,\dee\mainmeas\right)^{1/4}\right]\\
\leq& \int_E\exp\left[\sqrt{2}\left(\left(\varphi+\frac{9}{2}\right)^2\right)^{1/4}\right]\dee\mainmeas\\
=&\int_Ee^{\sqrt{2\varphi+9}}\dee\mainmeas\\
\leq&\int_Ee^{\sqrt{2\varphi}+3}\dee\mainmeas.
\end{aligned}\]
In other words,
$$\sqrt{2}\left(\int_E\varphi^2\, \dee\mainmeas\right)^{1/4}\leq 3+\log\int_Ee^{\sqrt{2\varphi}}\,\dee\mainmeas$$
and so
\begin{equation}\label{alt1}
2\left(\int_E\varphi^2\, \dee\mainmeas\right)^{1/2}\leq \left(3+\log\int_Ee^{\sqrt{2\varphi}}\,\dee\mainmeas\right)^2.
\end{equation}

Plugging this into (\ref{3}), we conclude that
\begin{equation}\label{first_part}
\int_{\lbrace u\leq 4\rbrace}\varphi|u|\,\dee\mainmeas\leq \left(3+\log\int_Ee^{\sqrt{2\varphi}}\,\dee\mainmeas\right)^2 \kl{\estmeas}{\mainmeas}^{1/2}.
\end{equation}

\paragraph{Second part of the proof.}

Instead of following the logic of the second part of the proof of \citet[Theorem 1]{Bolley:2005}, which fails to provide the result we are seeking, we can note the following:
\[
\int_{u>4}\varphi u\,\dee\mainmeas &\leq \frac{1}{(\log(5)-1)^{1/2}}\int_{u>4}\varphi(u+1)\left(\log(u+1)-1\right)^{1/2}\,\dee\mainmeas\nonumber\\
&\leq 2\left(\int_{u>4}\varphi^2(u+1)\,\dee\mainmeas\right)^{1/2}\left(\int_{u>4}\left[(u+1)\left(\log(u+1)-1\right)+1\right]\dee\mainmeas\right)^{1/2}\nonumber\\
&=2\left(\int_{u>4}\varphi^2\,\dee\estmeas\right)^{1/2}\left(\int_{u>4}h(u)\,\dee\mainmeas\right)^{1/2}\nonumber\\
&\leq2\left(\int_{E}\varphi^2\,\dee\estmeas\right)^{1/2}\kl{\estmeas}{\mainmeas}^{1/2}.\label{second}
\]
Now, since the function $t\mapsto \exp\left(\frac{\sqrt{2}\delta}{1+\delta}t^{1/4}\right)$ in increasing and convex on $\left[\frac{81(1+\delta)^4}{4\delta^4},+\infty\right)$, we can write
\[\begin{aligned}
&\exp\left[\frac{\sqrt{2}\delta}{1+\delta}\left(\int_E\varphi^2\, \dee\estmeas\right)^{1/4}\right]\\
\leq&\exp\left[\frac{\sqrt{2}\delta}{1+\delta}\left(\int_E\left(\varphi+\frac{9(1+\delta)^2}{2\delta^2}\right)^2\,\dee\estmeas\right)^{1/4}\right]\\
\leq& \int_E\exp\left[\frac{\sqrt{2}\delta}{1+\delta}\left(\left(\varphi+\frac{9(1+\delta)^2}{2\delta^2}\right)^2\right)^{1/4}\right]\dee\estmeas\\
=&\int_Ee^{\sqrt{2\delta^2\varphi/(1+\delta^2)+9}}\dee\estmeas\\
\leq&\int_Ee^{\delta\sqrt{2\varphi}/(1+\delta)+3}\dee\estmeas.
\end{aligned}\]
In other words,
$$\frac{\sqrt{2}\delta}{1+\delta}\left(\int_E\varphi^2\, \dee\estmeas\right)^{1/4}\leq 3+\log\int_Ee^{\delta\sqrt{2\varphi}/(1+\delta)}\,\dee\estmeas$$
and so
\begin{equation}\label{alt1}
2\left(\int_E\varphi^2\, \dee\estmeas\right)^{1/2}\leq \left(\frac{1+\delta}{\delta}\right)^2\left(3+\log\int_Ee^{\delta\sqrt{2\varphi}/(1+\delta)}\,\dee\estmeas\right)^2.
\end{equation}
Moreover, using H\"older's inequality,
$$\int_Ee^{\delta\sqrt{2\varphi}/(1+\delta)}\,\dee\estmeas\leq \left(\int_Ee^{\sqrt{2\varphi}}\,\dee\mainmeas\right)^{\delta/(1+\delta)}\left(\int_Ef^{1+\delta}\dee\mainmeas\right)^{1/(1+\delta)}$$
and so
\[\begin{aligned}
\int_{u>4}\varphi u\,\dee\mainmeas&\leq \left(\frac{1+\delta}{\delta}\right)^2\left(3+\frac{\delta}{1+\delta}\log \int_Ee^{\sqrt{2\varphi}}\,\dee\mainmeas+\frac{1}{1+\delta}\log \int_Ef^{1+\delta}\dee\mainmeas\right)^2\kl{\estmeas}{\mainmeas}^{1/2}\\
&\leq \left(27\left(\frac{1+\delta}{\delta}\right)^2+3\left(\log \int_Ee^{\sqrt{2\varphi}}\,\dee\mainmeas\right)^2+3\left(D_{1+\delta}(\estmeas|\mainmeas)\right)^2\right)\kl{\estmeas}{\mainmeas}^{1/2}
\end{aligned}\]
Combining this with (\ref{first_part}), we obtain the required result.

\subsection{Proof of \cref{thm:polynomial-moments}}

We have the following more general result, which we prove in the next section:

\begin{theorem}\label{thm:TV-polynomial-moments}
Let  $\varphi$ be a nonnegative measurable function on E and let $q,q'>1$ be such that $\frac{1}{q}+\frac{1}{q'}=1$. Then we have
$$\|\varphi(\estmeas-\mainmeas)\|_{TV}\leq \left[2\left(\int_E\varphi^2\,\dee\mainmeas\right)^{1/2}+2\left(\frac{1}{q}\int_E\varphi^{2q}\dee\mainmeas+\frac{1}{q'}\exp\left((q'-1)D_{q'}(\estmeas|\mainmeas)\right)\right)^{1/2}\right]\kl{\estmeas}{\mainmeas}^{1/2}.$$
\end{theorem}

As described in more detail in \cref{sec:proof-of-sqrt-EI}, \cref{thm:polynomial-moments} follows immediately from \cref{thm:TV-polynomial-moments}
 when we use $\varphi=\frac{1}{2}m(\paramrv',\cdot)$.

\subsection{Proof of \cref{thm:TV-polynomial-moments}}

We again assume, without loss of generality, that $\estmeas$ is absolutely continuous with respect to $\mainmeas$, with density $f$. We set $u:=f-1$ so that
$$\estmeas=(1+u)\mainmeas$$
and note that $u\geq -1$ and $\int_Eu\dee\mainmeas=0$. We also define
$$h(v):=(1+v)\log(1+v)-v,\quad v\in[-1,+\infty)$$
so that
\[
\kl{\estmeas}{\mainmeas}=\int_Eh(u)\dee\mainmeas.
\]
We note that $h\geq 0$. We split the total variation in the following way:
\begin{equation}\label{second_app1}
\int\varphi \,d|\estmeas-\mainmeas|=\int\varphi|u|\,\dee\mainmeas=\int_{\lbrace -1\leq u\leq 4\rbrace} \varphi|u|\,\dee\mainmeas+\int_{\lbrace u>4\rbrace} \varphi u\,\dee\mainmeas.
\end{equation}
Using (\ref{3}), we have that
\begin{equation}\label{second_app2}
\int_{\lbrace u\leq 4\rbrace}\varphi|u|\,\dee\mainmeas\leq 2\left(\int_E\varphi^2\,\dee\mainmeas\right)^{1/2} \kl{\estmeas}{\mainmeas}^{1/2}.
\end{equation}
Furthermore, using (\ref{second}), we have
\begin{equation}\label{second_app3}
\int_{u>4}\varphi u \,\dee\mainmeas\leq 2\left(\int_E\varphi^2\,\dee\estmeas\right)^{1/2}\kl{\estmeas}{\mainmeas}^{1/2}.
\end{equation}
Using Young's inequality, we obtain
$$\int_E\varphi^2\,\dee\estmeas=\int_E\varphi^2f\,\dee\mainmeas\leq \frac{1}{q}\int_E\varphi^{2q}\dee\mainmeas+\frac{1}{q'}\int_Ef^{q'}\dee\mainmeas=\frac{1}{q}\int_E\varphi^{2q}\dee\mainmeas+\frac{1}{q'}\exp\left((q'-1)D_{q'}(\estmeas|\mainmeas)\right),$$
which, together with (\ref{second_app1}), (\ref{second_app2}) and (\ref{second_app3}) gives the assertion.

\subsection{Proof of \cref{thm:polynomial-moments-simple}}

As described in more detail in \cref{sec:proof-of-sqrt-EI}, \cref{thm:polynomial-moments-simple} follows immediately from the following result:s
\begin{theorem}\label{thm:TV-polynomial-moments-simple}
Let  $\varphi$ be a nonnegative measurable function on E and suppose that $\estmeas$ and $\mainmeas$ are probability measures and $\estmeas\ll\mainmeas$. Then
$$\|\varphi(\estmeas-\mainmeas)\|_{TV}\leq\left(\int\varphi^2\,d\mainmeas\right)^{1/2}\left(\exp\left\{D_2(\estmeas|\mainmeas)\right\} - 1\right)^{1/2}.$$
\end{theorem}
\bprf
Let $f=\frac{\dee\estmeas}{\dee\mainmeas}$. We set $u:=f-1$ so that
$$\estmeas=(1+u)\mainmeas.$$
Note that the total variation can be expressed in the following way
\[\begin{aligned}
\int \varphi\,d|\estmeas-\mainmeas|=&\int\varphi |u|\, \dee\mainmeas\\
\leq& \left(\int\varphi^2\,\dee\mainmeas\right)^{1/2}\left(\int u^2\,\dee\mainmeas\right)^{1/2}\\
\leq& \left(\int\varphi^2\,\dee\mainmeas\right)^{1/2}\left(\int (f^2-2f+1)\,\dee\mainmeas\right)^{1/2}\\
=&\left(\int\varphi^2\,\dee\mainmeas\right)^{1/2}\left(\int f^2\,\dee\mainmeas-1\right)^{1/2}\\
=&\left(\int\varphi^2\,\dee\mainmeas\right)^{1/2}\left(\exp\left\{D_2(\estmeas|\mainmeas)\right\} - 1\right)^{1/2}.
\end{aligned}\]
\eprf